\pdfoutput=1
\documentclass[10pt,journal,compsoc]{IEEEtran}

\ifCLASSOPTIONcompsoc
  \usepackage[nocompress]{cite}
\else
  \usepackage{cite}
\fi

\ifCLASSINFOpdf
  \usepackage[pdftex]{graphicx}
  \graphicspath{{../pdf/}{../jpeg/}}
  \DeclareGraphicsExtensions{.pdf,.jpeg,.png}
\else
  \usepackage[dvips]{graphicx}
  \graphicspath{{../eps/}}
  \DeclareGraphicsExtensions{.eps}
\fi

\usepackage{amsmath}
\newtheorem{myDef}{DEFINITION}[section]\newtheorem{myPro}{PROOF}\newtheorem{myclaim}{CLAIM}

\usepackage{fixltx2e}
\usepackage{stfloats}

\ifCLASSOPTIONcaptionsoff
  \usepackage[nomarkers]{endfloat}
 \let\MYoriglatexcaption\caption
 \renewcommand{\caption}[2][\relax]{\MYoriglatexcaption[#2]{#2}}
\fi

\usepackage{url}

\usepackage{amsfonts}
\usepackage{booktabs}
\usepackage{multirow}
\usepackage{makecell}
\usepackage{bm}

\usepackage{algorithm}
\usepackage{algorithmicx}
\usepackage{array}
\usepackage{algpseudocode}

\usepackage{enumerate}
\usepackage{eqparbox}

\usepackage{multirow}
\usepackage{booktabs}

\usepackage{array,graphicx,subfig}

\begin{document}

\title{Hierarchical Adaptive Pooling by Capturing High-order Dependency for \\Graph Representation Learning}

\author{Ning Liu, Songlei Jian, Dongsheng Li, Yiming Zhang, Zhiquan Lai and Hongzuo Xu
\IEEEcompsocitemizethanks{\IEEEcompsocthanksitem Ning Liu, Songlei Jian, Dongsheng Li, Yiming Zhang, Zhiquan Lai and Hongzuo Xu are with the College of Computer, National University of Defense Technology, China. \protect\\
E-mail: {liuning17a, jiansonglei, dsli, ymzhang, zqlai, xuhongzuo13}\\@nudt.edu.cn
}
\thanks{Manuscript received December 19, 2020; revised April 26, 2021. (Corresponding author: Dongsheng Li.)}}


\markboth{IEEE Transactions of knowledge and Data Engineering, ~Vol.~X, No.~X, February~2021}%
{Shell \MakeLowercase{\textit{et al.}}: Bare Demo of IEEEtran.cls for Computer Society Journals}

\IEEEtitleabstractindextext{%
\begin{abstract}
Graph neural networks (GNN) have been proven to be mature enough for handling graph-structured data on node-level graph representation learning tasks. However, the graph pooling technique for learning expressive graph-level representation
is critical yet still challenging. Existing pooling methods either struggle to capture the local substructure or fail to effectively utilize high-order dependency, thus diminishing the expression capability. In this paper we propose HAP, a hierarchical graph-level representation learning framework, which is adaptively sensitive to graph structures, i.e., HAP clusters local substructures incorporating with high-order dependencies. HAP utilizes a novel cross-level attention mechanism MOA to naturally focus more on close neighborhood while effectively capture higher-order dependency that may contain crucial information. It also learns a global graph content GCont that extracts the graph pattern properties to make the pre- and post-coarsening graph content maintain stable, thus providing global guidance in graph coarsening. This novel innovation also facilitates generalization across graphs with the same form of features. Extensive experiments on fourteen datasets show that HAP significantly outperforms twelve popular graph pooling methods on graph classification task with an maximum accuracy improvement of 22.79\%, and exceeds the performance of state-of-the-art graph matching and graph similarity learning algorithms by over 3.5\% and 16.7\%.
\end{abstract}

}

\maketitle

\IEEEdisplaynontitleabstractindextext

\IEEEpeerreviewmaketitle

\IEEEraisesectionheading{\section{Introduction}\label{sec:introduction}}

\IEEEPARstart{A}{lthough} data that can be represented as grid structure on Euclidean domains, such as images~\cite{zhang2017beyond}, video~\cite{fan2016video}, speech~\cite{palaz2015analysis}, and texts~\cite{yin2017comparative}, has closely connections with daily life, there is another major category of Non-Euclidean data, namely graph, which is constructed by irregularly-arranged nodes and the connection-indicated edges. Examples include social networks~\cite{myers2014information}, citation networks~\cite{shibata2012link}, road networks~\cite{hu2019stochastic} and bioinformatics~\cite{li2019deep}. Different from Euclidean data, convolution and pooling operation in Convolutional Neural Networks (CNNs) cannot be directly applied to Non-Euclidean graph-structured data due to the irregularity and nondeterminacy of the neighborhood for the central node. Consequently, it is important to learn a sufficiently expressive representation for graph-structured data in a reasonable way.

\subsection{Motivation}
A great deal of research on Graph Neural Networks (GNNs) has emerged to generalize the great success of convolution in CNNs to graph-structured data. In GNNs, convolution operation is evolved into neighborhood message aggregation of the central node along edges, thus capturing both node features and graph structural information. Following this principle, various GNNs have been proposed, such as GCN~\cite{kipf2016semi-supervised}, graphSAGE~\cite{hamilton2017inductive} and GAT~\cite{velivckovic2017graph}. All of them have achieved significant prosperity for graph representation learning tasks, especially for node-level representation based tasks, including node classification~\cite{KipfSemi,velivckovic2017graph} and link prediction~\cite{schlichtkrull2018modeling,vashishth2019composition}. However, as for graph-level representation learning tasks, such as graph classification~\cite{bruna2013spectral, Ying2018Hierarchical,gao2019graph}, graph matching~\cite{li2019graph} and graph similarity learning~\cite{bai2019simgnn}, convolution operation alone is deficient. It is nontrivial to potentially empower GNNs to produce discriminative graph-level representations with the help of pooling operation.

To remedy this problem, a few researchers have tried to further generalize the pooling mechanism from CNNs to GNNs for graph-level representation learning. Accordingly, it is natural to raise the question: what are the basic criteria for a high-quality pooling method in GNNs? Actually, different from the pooling operation in CNNs for reducing the number of computational parameters and preserving the invariance, the basic idea of graph pooling techniques is a node feature aggregator throughout the entire graph, analogous to the neighborhood aggregator in graph convolution. Therefore, a good graph pooling 
method should encourage graphs with approximate topology and similar node features to have resemblant representations to some extent. As a result, the major challenge for a high-quality pooling mechanism is to define a method which both effectively maintains pivotal node features and explicitly captures important structural information. To address this challenge, previous works have proposed graph pooling architectures in three ways.

 First, universal maximum/average pooling methods~\cite{li2016gated, bai2019simgnn} are intuitively extended to graph models by simple element-wise max- or mean-downsampling through all node features. But such methods have been proved to ignore feature multiplicities, as well as completely miss the structural information~\cite{xu2018powerful}. Graphs with different corresponding structures may get the same representation. 
 
 Second, \emph{Top-K} methods~\cite{vinyals2015order, gao2019graph, zhang2018end, huang2019attpool, lee2019self} sort graph nodes in a consistent order with scores on behalf of the importance, only \emph{K} nodes with the highest scores are selected to form the pooled graph. As a result, sets of nodes except for the top \emph{K} ones are definitely discarded, which may involve important features. Meanwhile, very few \emph{Top-K} methods utilize the local substructure information for scoring stage, thus resulting a strong possibility of the selected nodes to be isolated, which may influence the information propagation in subsequent GNN layers.
 
\begin{figure}[!t]
\centering
\includegraphics[width=1\columnwidth]{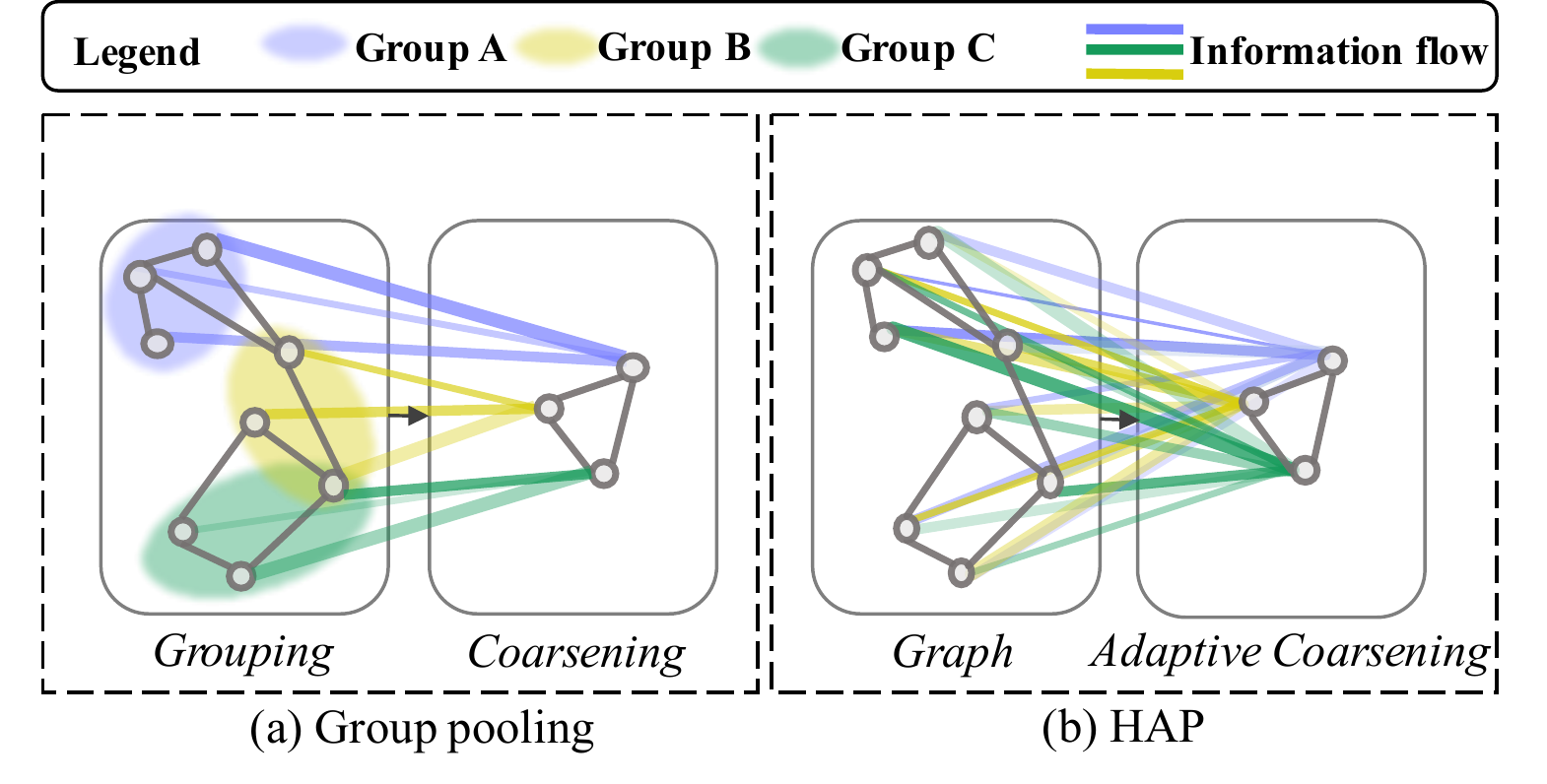}
\caption{Differences of receptive field during clustering between group pooling methods and HAP.}
\label{fig:difference}
\end{figure}

 Third, advanced graph pooling methods, such as DiffPool~\cite{Ying2018Hierarchical} and ASAP~\cite{ranjan2020asap}, learn graph representations in a hierarchical grouping manner for capturing more comprehensive local substructures widespread in graphs. They usually group nodes into smaller clusters with multiple levels and utilize the final level as the graph representation. However, grouping operation is usually executed on a fixed 1-hop neighborhood, thus forcing the information to flow from a certain neighborhood to the specific coarsened cluster and neglecting the higher-order dependency among nodes that may hold significant information, as shown in \figurename~\ref{fig:difference}(a).
 

\subsection{Contributions}
To the best of our knowledge, no existing graph pooling methods adaptively handle the graph local substructures and high-order dependency, while capturing node features. And there is a general lack of systematical consideration to the generalization of graph-level representations affected by pooling methods. To bridge the above gap, we propose a novel hierarchical graph pooling framework called \textbf{\underline{H}ierarchical \underline{A}daptive \underline{P}ooing} (\textbf{HAP}). The main contributations can be summarized as follows.

\begin{itemize}
\item We intruduce \textbf{HAP}, a supervised hierarchical pooling framework. HAP is capable of preserving node feature information with adaptive graph structure sensitivity to both local substructures and high-order dependency. We provide a more comprehensible framework by offering exhaustive theoretical analysis for computational complexity, permutation invariance and the design validity of HAP.

\item We propose \textbf{master-orthogonal attention (MOA)}, a novel cross-level attention mechanism specifically designed for hierarchical graph pooling. MOA can be leveraged to capture cross-level interactions under the guidance of graph pattern properties in a more efficient and effective way. MOA also acts as a soft substructure extractor. Attention weights for nodes in the receptive field of a possible local cluster are much higher than those beyond the receptive field. This ensures the local-substructure sensitivity and introduces high-order node features to it.

\item We design \textbf{GCont}, an auto-learned global graph content playing a significant role in MOA. The key innovation is that it incorporates high-level global pattern properties into pooling method, making MOA sensitive to the latent graph characteristics and produce a more adequate graph-level representation without interference of artificial factors. It is relatively stable during hierarchical pooling and flexible enough to be learned heuristically. GCont also guarantees the generalization ability across graphs with the same form of features.
\end{itemize}

Extensive experiments demonstrate that (1) HAP significantly outperforms twelve graph pooling methods on six real-world datasets for graph classification task with a maximum accuracy improvement of 22.79\%; (2) HAP sharply outperforms the state-of-the-art GMN~\cite{li2019graph}, which is specifically designed for graph matching task, by boosting the accuracy up to 3.5\%; (3) HAP also achieves a maximum accuracy gain of 16.7\% comparing with conventional approximate GED algorithms and GNN-based graph similarity learning models; (4) The graph coarsening module in HAP dramatically enhances the expression ability of existing graph pooling architecture for graph-level tasks; (5) HAP achieves good generalization ability across graphs with the same form of features; and (6) HAP provides meaningful visualization of graph-level representations.


\section{Related Work}
In this section, we summarize some related works on graph pooling in GNNs covering the two main types of methods: supervised and unsupervised.
\subsection{Supervised Pooling}
Supervised pooling methods can be divided into flat pooling and hierarchical pooling according to whether the graph-level representation is aggregated in a flat or hierarchical way with a view to local substructures. Further, flat pooling methods also cover two families: universal pooling and \emph{TopK} pooling, depending on the number of nodes participated in the final aggregation.
\subsubsection{Flat Universal Pooling}
Flat universal pooling methods take all the nodes into consideration. Earlier works directly learn from CNNs to use mean- or max-pooling method to extract features. Subsequently, Xu et al.~\cite{xu2018powerful} find that sum-pooling is much more powerful because no matter mean or max aggregator ignores the multiformity of features, thus struggling in distinguishing graphs with nodes that have repeating features. Some other works rely on content-based attention operation. In Gated Graph Neural Networks (GG-NNs)~\cite{li2016gated}, the graph-level output is defined by a soft attention mechanism for each node to decide which is more relevant to the graph-level task. Message Passing Neural Networks (MPNNs)~\cite{gilmer2017neural} further utilizes Set2Set~\cite{vinyals2015order} method to take the order of nodes into consideration and find the importance of each node to the graph-level representation through time-consuming iterative soft-attention. In SimGNN~\cite{bai2019simgnn} and UGRAPHEMB~\cite{bai2019unsupervised}, a graph content is defined as the average of node features and the attention is executed between nodes and it. Obviously, such man-made design makes the final graph-level representation infinitely close to the output of mean-pooling method, which is an inefficient method mentioned above.
\subsubsection{Flat Top-K pooling}
Flat Top-K pooling methods score the nodes according to the importance. Nodes with \emph{k}-largest scores are preserved to form the new graph. SortPooling~\cite{zhang2018end} method refers to the graph label method WL~\cite{shervashidze2011weisfeiler-lehman}, it regards the output node features of each GCN layer as the continuous WL colors and sorts the nodes according to the last GCN layer's colors. AttPool~\cite{huang2019attpool} calculates the scores using a global soft-attention mechanism. Further more, a local attention method accesses node degree information, which contributes to keep a balance between the importance and the dispersion. gPool~\cite{gao2019graph} develops new ideas that use the projection of node features to a trainable projection vector as node score. SAGPool~\cite{lee2019self} considers both node features and graph topology during pooling by taking GCN to calculate attention scores. However, important information existed in the abandoned nodes may be ignored, which should be explicitly captured in graph pooling. Moreover, no structural relationships among nodes are acquired during pooling, thus may lead to unconnectedness of the selected nodes.

\subsubsection{Hierarchical Group Pooling}
Due to the fact that local substructures are present in real-world graphs, hierarchical group pooling methods come into being. DiffPool~\cite{Ying2018Hierarchical} is the first differentiable group pooling approach that learns a dense assignment matrix to group 1-hop neighborhood of nodes into clusters in each hierarchical layer. Subsequently, to address the sparsity concerns in DiffPool, ASAP is proposed, which combines both TopK and group methods. Clusters are generated by aggregating \emph{h}-hop neighbors of each central node to leverage the graph structure, then only top scoring clusters are maintained. Actually, ASAP~\cite{ranjan2020asap} still cannot guarantee the connectivity between the selected clusters. Actually, Mesquita et al. in~\cite{mesquita2020rethinking} indicate that a successful graph pooling method should not be restricted to nearby nodes. Thus we propose that high-order structural dependency also contributes to the construction of a good pooling region. 

\subsection{Unsupervised Pooling}
Loss functions of the aforementioned graph pooling methods are usually task-based supervised except for DiffPool that exploits a link prediction loss and enforces nearby nodes to be pooled together. Recently, there is a growing interest in unsupervised graph pooling by minimizing the objective related to graph structure characteristic or borrowed from graph theory. StructPool~\cite{yuan2019structpool} employs conditional random fields (CRFs) to capture high-order structural relationships by minimizing the Gibbs energy. MinCutPool~\cite{bianchi2020spectral} continuously relaxes the normalized minCUT problem in graph theory and optimizes the cluster assignments by minimizing this objective. UGRAPHEMB~\cite{bai2019unsupervised} utilizes well-accepted and domain-agnostic graph proximity metrics to provide extra graph-graph proximity guidance during learning. These novel ideas offer the possibility of breaking a logjam of current graph pooling research. 

Something also worth mentioning is that there is a common challenge for no matter universal pooling, \emph{Top-K} pooling or group pooling, i.e., the element-wise aggregation, score ranking and cluster assignment learning processes are merely executed on a single fixed graph, lacking the inductive capability for entirely new graphs.

\section{Preliminaries}

\subsection{Problem Statement}
A graph is represented as $G=\left(\mathcal{V}, \mathcal{E}, \mathcal{X}\right)$, where $\mathcal{V}=\left\{\mathbf{v}_{i}\right\}_{i=\left\{1,\cdots,N\right\}}$ denotes the set of nodes, $\mathbf{e}_{ij}=\left(\mathbf{v}_{i}, \mathbf{v}_j\right)\in \mathcal{E}$ is the edge link between node $\mathbf{v}_{i}$ and $\mathbf{v}_{j}$, and $\mathcal{X}=\left\{\mathbf{x}_{1}, \mathbf{x}_{2},\cdots,\mathbf{x}_{N}\right\}$ consists the node labels (no node labels are provided in some cases). For a graph $G$ with $N=\left|\mathcal{V}\right|$ nodes and $\left|\mathcal{E}\right|$ edges, $\mathbf{A} \in \mathbb{R}^{N \times N}$ represents the weighted adjacency matrix and $\mathbf{D} \in \mathbb{R}^{N \times N}$ is a diagonal matrix that diagonal elements stand for the degree of nodes. $\mathbf{H} \in \mathbb{R}^{N \times F}$ denotes the node feature matrix and $\mathbf{h}_{G} \in \mathbb{R}^{F_{G}}$ is the graph-level embedding. A label $Y$ may also be attached to the graph $G$. Detailed notations are summarized in \tablename ~\ref{tab:notation}. Given a graph dataset, the graph pooling task aims to learn a mapping $f : \mathbf{H} \rightarrow \mathbf{h}_{G}$ from a node feature matrix to a single graph representation.

\begin{table}[!t]
  \caption{Notations}
  \label{tab:notation}
  \begin{tabular}{c|c}
    \toprule
     \textbf{Notations} & \textbf{Definitions or Descriptions} \\
    \midrule
    $G$, $G'$     & the input/coarsened graph                              \\
$\mathcal{V}$, $\mathcal{E}$, $\mathcal{X}$   & the node/edge/node label set of $G$                     \\
$N$, $N'$     & the number of nodes of $G$/$G'$                            \\
$\mathbf{A}$,  $\mathbf{A}'$    & the adjacent matrix of $G$/$G'$                            \\
$\mathbf{D}$         & the degree matrix of $G$                               \\
$\mathbf{H}$, $\mathbf{H}'$    & the node feature matrix of $G$/$G'$             \\
$\mathbf{h}_{G}$ 	& the graph-level embedding of $G$ \\
$K$	& the number of graph coarsening modules\\
$\mathbf{C}$         & the auto-learned global graph content                  \\
$\mathbf{C}_{(i, \cdot)}$      & row in C refers to a node of the source graph $G$      \\
$\mathbf{C}_{(\cdot, j)}$      & column in C refer to a cluster of the target graph $G$'   \\
$\mathbf{M}$         & the MOA matrix                        \\
$F$        &   the dimension of input node feature \\
$F'$		& the dimension of the output node feature\\
$F_{G}$		&the dimension of the graph level embedding\\
$Y$ & the label of graph $G$ \\

    \bottomrule
  \end{tabular}
\end{table}

\subsection{Downstream Tasks}
Most previous works only focus on the application of graph pooling for graph classification, which is an important graph-level representation learning task but partial. In this paper, we formally summarize and define the downstream graph pooling tasks and conduct exhaustive experiments over them:

\begin{itemize}
    \item \textbf{Graph Classification:} Given an input graph $G$, the graph classification task tries to learn a mapping from the graph to the corresponding graph label.
    \item \textbf{Graph Matching:} Given an input graph pair $\left(G_{1},G_{2}\right)$, the graph matching task aims to determine whether $G_{1}$ and $G_{2}$ are isomorphic\footnote{For a pair of graph, graph isomorphism decides whether there exists a bijective function between them so that nodes are connected in the same way.} to each other.
    \item \textbf{Graph Similarity Learning:} Given an input graph triple $\left(G_{1},G_{2},G_{3}\right)$, the graph similarity learning task manages to explore whether $G_{1}$ is much closer to $G_{2}$ or $G_{3}$.
\end{itemize}

\subsection{Graph Neural Networks}
Given node features $\mathbf{H}=\left\{{h}_{i},\cdots, {h}_{N}\right\}$ and graph structure $\mathbf{A}$, modern GNNs usually learn useful node representations in an neighborhood aggregation fashion following general ``message-passing'' architecture. The forward process comprises two phases, each of which iteratively runs for $\mathbf{L}$ time steps. The message passing phase aggregates information along edges of the central node from its neighbors. Then the combination phase updates the representation of the central node based on the message:
\begin{equation}\label{equ:GNNs-message}
	\mathbf{MESSAGE}_{\mathcal{V}\rightarrow i}^{\left(l\right)} = \mathbf{AGGREGATE}^{\left(l\right)}\left({h}_{j}^{\left(l-1\right)}: \left(j, i\right)\in \mathcal{E}\right)
\end{equation}

\begin{equation}\label{equ:GNNs-combine}
	{h}_{i}^{\left(l\right)}=\mathbf{COMBINE}^{\left(l\right)}\left({h}_{i}^{\left(l-1\right)},\mathbf{MESSAGE}_{\mathcal{V}\rightarrow i}^{\left(l\right)}\right)
\end{equation}
where ${h}_{i}^{\left(l\right)}$ is the embedding of node $i$ at the $l$-th iteration that is initialized as ${h}_{i}^{\left(0\right)}={h}_{i}$, and ${h}_{j}^{\left(l-1\right)}$ is the node feature vector of node $i$'s neighbor depending on the adjacency matrix.

There are multiple selectable implementations of $\mathbf{AGGREGATE}^{\left(l\right)}\left(\cdot\right)$ and $\mathbf{COMBINE}^{\left(l\right)}\left(\cdot\right)$ adapted successfully to different GNN models. Actually, our HAP pooling framework can be consolidated into any GNN models following the implementation of Equation~\ref{equ:GNNs-message} and Equation~\ref{equ:GNNs-combine}. After $\mathbf{L}$ times iteration, the representation of the central node captures the features and structural information within its $\mathbf{L}$-hop neighborhood.

%


\subsection{Graph Attention}\label{sec:attention}

Graph attention mechanism executed between a query $q$ and a key $k$ allows for allocating diverse alignment scores $\alpha_{ij}$ to different parts of the input, making the model focus on the most relevant portion. Existing graph attention mechanism can be divided into \emph{node-level attention} and \emph{master-level attention} according to the attention scope. Specifically, node-level attention covers both self-attention and cross-attention.

\textbf{Hard-Self-Attention (HSA)}~\cite{velivckovic2017graph} chooses both $q$ and $k$ from the node features of the single input graph to find the node dependency on itself:
\begin{equation}\label{equ:hsa}
	\alpha_{ij}=softmax\left( \sigma\left(u^{\top}\left [ \mathbf{W}h_{i}\Vert \mathbf{W}h_{j}\right]\right) \right), \left\{h_{i}\right\}_{i \in \mathcal{V}_{1}},\left\{h_{j}\right\}_{j \in \mathcal{V}_{1}}
\end{equation}	
where $u^{\top}$ and $\mathbf{W}$ are trainable parameters, and $[\cdot \Vert \cdot]$ is a concatenation operation.

\textbf{Soft-Self-Attention (SSA)}~\cite{li2016gated} decides which nodes are relevant to the current graph-level task, so that $q$ is defined as node feature but no specific key $k$ is provided:
\begin{equation}\label{equ:ssa}
	\alpha_{ij}=softmax\left( \sigma\left(u^{\top}\left ({\mathbf{W}}h_{j}\right)\right) \right), \left\{h_{j}\right\}_{j \in \mathcal{V}_{1}}\end{equation}

\textbf{Cross-Attention (CA)}~\cite{li2019graph} captures the differences between graphs by doing comparisons across the pair of graphs through choosing $q$ and $k$ from the node features of pairwise input, thus fusing information from both graphs:
\begin{equation}\label{equ:ca}
	\alpha_{ij}=softmax\left( \sigma\left(u^{\top}\left [ {\mathbf{W}}h_{i}\Vert {\mathbf{W}}h_{j}\right]\right) \right), \left\{h_{i}\right\}_{i \in \mathcal{V}_{1}},\left\{h_{j}\right\}_{j \in \mathcal{V}_{2}}
\end{equation}

\textbf{Master-Attention (MA)}~\cite{ranjan2020asap, bai2019simgnn, bai2019unsupervised} concentrates on the interaction between nodes and the master they belong to, so that $q$ and $k$ denote node feature and master feature separately. The master function is generally defined as sum or max operation of the constituent nodes:
\begin{equation}\label{equ:ma}
	\alpha_{ij}=softmax\left( \sigma\left(u^{\top}\left [ {\mathbf{W}}\cdot\mathbf{Master}\left(h_{j}\right)\Vert {\mathbf{W}}h_{j}\right]\right) \right), \left\{h_{j}\right\}_{j \in \mathcal{V}_{1}}
\end{equation}
\begin{equation}\label{equ:master}
	\mathbf{Master}= \sum_{v_{j} \in c_{i}\left(v_{j}\right)}{\left(h_{j}\right)}
\end{equation}

\section{The Proposed Method: HAP}
In this section, we present HAP, a hierarchical graph pooling framework for graph-level tasks. Its key idea is the graph coarsening module supported by novel graph pattern property extracting technique GCont and cross-level attention mechanism MOA complementing and reinforcing each other, which not only prompts the GNN model to be sensitive to both local substructures and high-order dependency, but also empowers it with stronger generalization ability. Below, we discuss the components of HAP in details.

\subsection{Hierarchical Framework}
\begin{figure*}[t]
\centering
\includegraphics[width=0.9\textwidth]{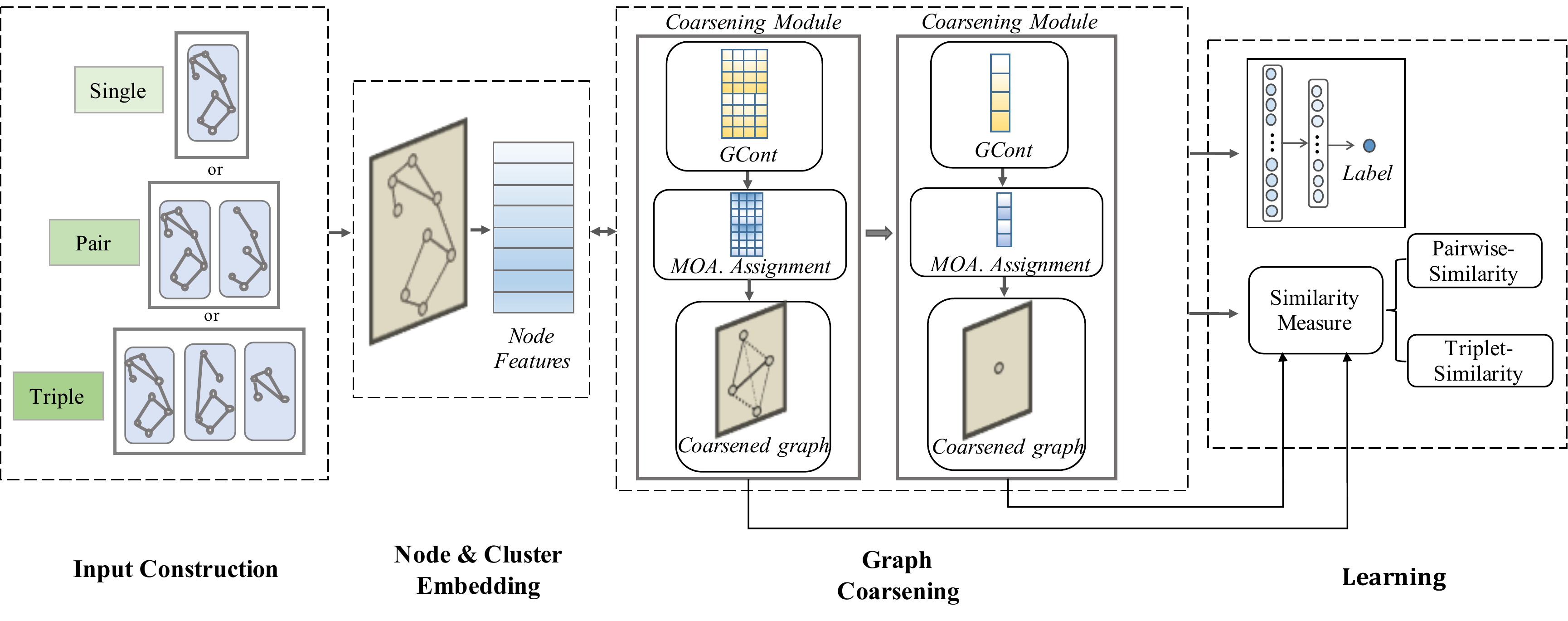}
\caption{The overview framework of HAP which conducts graph coarsening with GCont and MOA mechanism, and compares graphs with the hierarchical similarity measure. }
\label{fig:model structure}
\end{figure*}

Figure~\ref{fig:model structure} illustrates the overall architecture of the HAP. Given single, pairwise or triplet input graphs for differentiated graph-level tasks, HAP extracts the node features and graph structure information for an end-to-end training. The process
can be decomposed into six main steps:

\begin{enumerate}[Step-1:]
	\item \textbf{Input Construction} Single input graph classification and pairwise graph matching task require no special operation on the given dataset, which consists of single or pairs of graphs with true labels to indicate which class the graph belongs to or whether the pair is matching or not. However, a triplet generator is necessary for graph similarity learning task.
	\item \textbf{Node \& Cluster Embedding} Subsequently, single or pairs of graphs, or the generated triplets are transferred into a node \& cluster embedding module to learn a low-dimensional node vector representation for each node or coarsened cluster. The cluster representations will be maintained for hierarchical similarity measuring.
	\item \textbf{Graph Coarsening-\uppercase\expandafter{\romannumeral1}} Then, a learnable GCont defines a coarsening preparation step for each graph by extracting global pattern properties. Rows and columns of it correspond to source nodes before coarsening and target clusters after coarsening, respectively.
	\item \textbf{Graph Coarsening-\uppercase\expandafter{\romannumeral2}} Furthermore, the MOA mechanism is utilized to obtain an attention assignment. The attention coefficient matrix, each element of which indicates the contribution of the node from the source graph $G$ to the cluster from the target graph $G'$, is derived from the GCont.
	\item \textbf{Graph Coarsening-\uppercase\expandafter{\romannumeral3}} Afterwards, a cluster formation function $\Omega: G \in \mathbb{R}^{N\times N}\rightarrow G'\in \mathbb{R}^{N'\times N'}$ is learned to compute the cluster representation after one coarsening. 
	\item \textbf{Learning:} Executes the loop between Step-2 and Step-5 until reaching a satisfied graph scale. HAP then calculates corresponding task loss with hierarchical graph representation to constantly optimize all the weight parameters.
\end{enumerate}

\subsection{Input Construction}\label{sec:triplet}
For graph similarity learning task, training and testing data in the form of triplet is essential to learn the relative similarity among graphs. Based on the best of our knowledge, there is no ready-to-use graph dataset in triplet form. To bridge the above gap, we propose a triplet generator in this subsection.

Given a dataset with single graphs, we denote it as $\mathcal{S}$, the similarity between every two graphs $G_{i}$ and $G_{j}$ can be measured under a graph-graph proximity metric $f(\cdot)$, such as Graph Edit Distance (GED). The smaller the GED, the more similar the pair. Then the pairwise ground-truth proximity is denoted as follows:

\begin{equation}
	\mathcal{P}_{GED}\leftarrow\big\{g_{ij}\big|g_{ij}=f(G_{i},G_{j}),\forall i,j \in |\mathcal{S}|\big\}
\end{equation}

Afterwards, we conduct triplets by fixing the first position with one graph and randomly choose two disparate graphs to fill the rest two positions:

\begin{equation}
	\mathcal{T}\leftarrow \big\{\gamma \big|\gamma=\left<G_{i},G_{j},G_{k} \right>,\forall i,j,k \in |\mathcal{S}|, j \neq k\big\}
\end{equation}

Synchronously, the ground truth triplet proximity is generated as follows, in which a positive number for the element $r_{ijk}$ means that graph $G_{i}$ is much similar to graph $G_{k}$ and a negative number means that graph $G_{i}$ is much similar to graph $G_{j}$:

\begin{equation}
	\mathcal{T}_{GED}\leftarrow\big\{r_{ijk}\big|r_{ijk}=g_{ij}-g_{ik},\forall i,j,k \in |\mathcal{S}|, j \neq k\big\}
\end{equation}

\subsection{Node \& Cluster Embedding}
There is a demand for node or cluster embedding to extract node or cluster features before going to the next graph coarsening module. In this paper, we choose to employ a two-layer GAT~\cite{velivckovic2017graph} or GCN~\cite{kipf2016semi-supervised} as basic components since they are all well capable of capturing the local structure information of a node. Actually, any mainstream GNNs can also be integrated into the HAP framework. And please note that the number of GAT or GCN layers depends on the real application graph data.

For the \emph{k}-th layer in GAT, it takes graph $G$'s adjacent matrix $\mathbf{A}_{k}$ and the hidden representation matrix $\mathbf{H}_{k}$ as input, then formulates the $\mathbf{AGGREGATE^{\left(k\right)}\left(\cdot\right)}$ phase in a weighted-attention-based operator:
\begin{equation}\label{eq:GAT}
    \mathbf{H}_{k+1}=\sigma\left(\mathbf{A}_{k}\mathbf{O}_{att}\mathbf{H}_{k}\mathbf{W}_{k}\right)
\end{equation}
where $\sigma\left(\cdot\right)$ is the non-linear activation function such as \emph{ReLU} or \emph{Sigmoid}, $\mathbf{O}_{att}$ is a trainable global attention assignment among all nodes, and $\mathbf{A}_{k}\mathbf{O}_{att}$ picks one-hop neighborhood attention. $\mathbf{W}_{k}$ is a trainable weight matrix.

Similarly, the implementation of Equation~\ref{equ:GNNs-message} and Equation~\ref{equ:GNNs-combine} for forward-propagation operation of GCN is defined as:

\begin{equation}\label{eq:GCN}
    \mathbf{H}_{k+1}=\sigma\left(\tilde{\mathbf{D}}_{k}^{-\frac{1}{2}}\tilde{\mathbf{A}}_{k}\tilde{\mathbf{D}}_{k}^{-\frac{1}{2}}\mathbf{H}_{k}\mathbf{W}_{k}\right)
\end{equation}
where $\tilde{\mathbf{A}}_{k}$ is the adjacent matrix plus self-connections (i.e., $\tilde{\mathbf{A}}_{k}=\mathbf{A}_{k}+\mathbf{I}$), $\tilde{\mathbf{D}}_{k}$ is the degree matrix of $\mathbf{A}_{k}$ (i.e., $\tilde{\mathbf{D}}=\sum_{j}{\tilde{\mathbf{A}}_{ij}}$), and $\tilde{\mathbf{D}}_{k}^{-\frac{1}{2}}\tilde{\mathbf{A}}_{k}\tilde{\mathbf{D}}_{k}^{-\frac{1}{2}}$ is the symmetric normalized Laplacian for graph $G$. With one convolutional layer, GCN is able to preserve the first-order neighborhood information between nodes. By stacking multiple GCN layers, it is capable to encode higher-order (e.g., k-hop) neighborhood relationships.

Specifically, different from classical GAT or GCN where graph scale is stable throughout the whole training, HAP scales down nodes into clusters in the graph coarsening module before transferring them to the next node \& cluster embedding layer. As a result, $\mathbf{A}_{k}$, $\tilde{\mathbf{D}}_{k}$ and $\tilde{\mathbf{A}}_{k}$ change with the action of graph coarsening (cf. Eq.~\ref{eq:A'}).

%

\subsection{Graph Coarsening}
\begin{figure*}[t]
\setlength{\abovecaptionskip}{0.cm}
\setlength{\belowcaptionskip}{-0.5cm}
\centering
\includegraphics[width=0.8\textwidth]{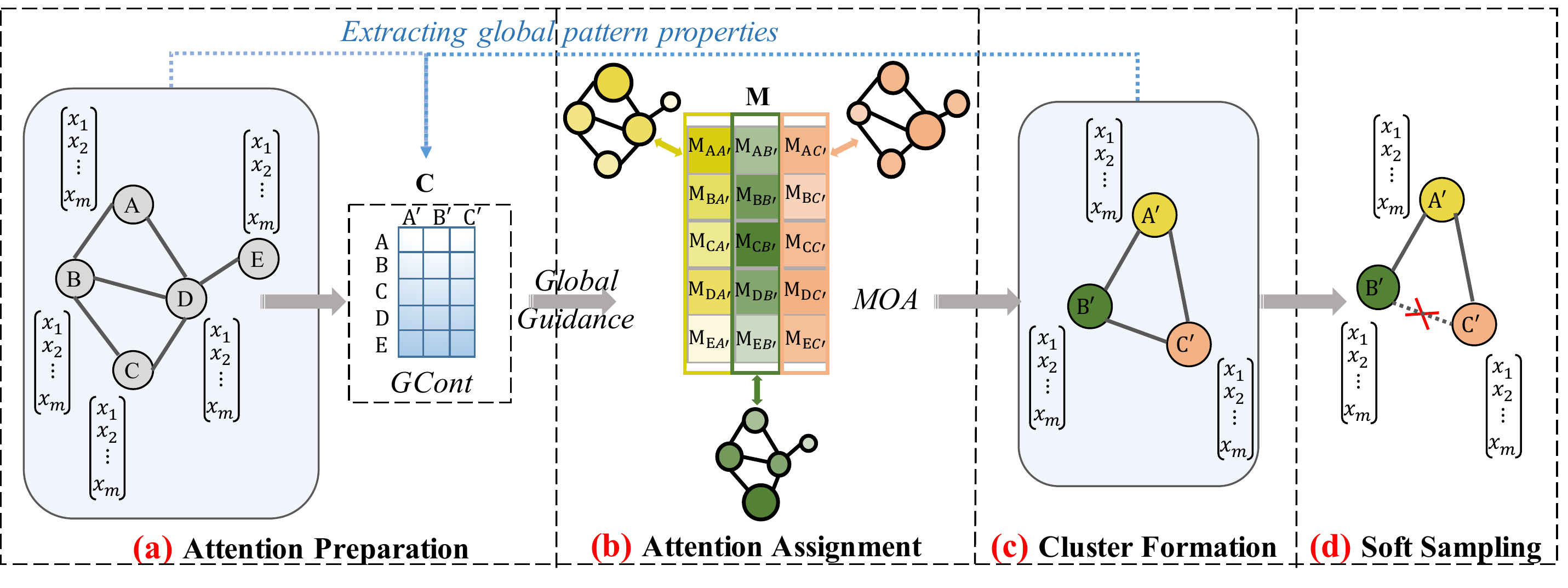}
\caption{An illustration of the graph coarsening module. The whole procedure contains four steps: (a) attention preparation by constructing GCont in which rows and columns stand for nodes from the source graph and the coarsened graph separately; (b) computing cross-level attention assignment by using MOA; (c) cluster formation by aggregating information from the source graph; (d) executing soft sampling for edges in the target coarsened graph.}
\label{fig:MOA module}
\end{figure*}

We achieve graph coarsening through graph global pattern property extracting technique GCont and cross-level attention mechanism MOA. We show the graph coarsening module architecture in \figurename~\ref{fig:MOA module} and elaborate the details in this subsection. Further, algorithm~\ref{alg:oamodule} gives the pseudocode for the graph coarsening module.

\subsubsection{Attention Preparation using GCont}
Given node features for the source graph, the task of coarsening process is to learn the cluster assignment matrix through attention mechanism. However, one important thing ignored by all the group pooling methods is that the pre- and post-coarsening graph content should remain stable without loss of important information. We observe that both DiffPool~\cite{Ying2018Hierarchical} and ASAP~\cite{ranjan2020asap} receive no global guidance. Hence, we propose \emph{GCont}, an auto-learned global graph content sustaining the coarsening process.

As an initial step, we propose using one learnable linear transformation, parametrized by the weight matrix $\mathbf{T} \in \mathbb{\mathbb{R}}^{F \times N'}$ to generate GCont. The simple linear transformer also combines scalability with the ability to deal with relatively larger graphs. The global graph content is converted from the node feature matrix $\mathbf{H}$ as:
\begin{small}
\begin{equation}\label{eq:global graph content}
\mathbf{C}=
\begin{bmatrix}
\sum_{i=1}^F\mathbf{H}_{1i}\mathbf{T}_{i1} &  \cdots   & \sum_{i=1}^F\mathbf{H}_{1i}\mathbf{T}_{iN'}  \\

\vdots & \vdots  & \ddots   & \vdots  \\
\sum_{i=1}^F\mathbf{H}_{Ni}\mathbf{T}_{i1} & \cdots\  & \sum_{i=1}^F\mathbf{H}_{Ni}\mathbf{T}_{iN'}  \\
\end{bmatrix}
\end{equation}
\end{small}
where $\mathbf{H}_{ij}$ and $\mathbf{T}_{ij}$ indicate element in the position of the $i$-th row and $j$-th column of matrix $\mathbf{H}$ and $\mathbf{T}$ separately. $\mathbf{C}\in \mathbb{R}^{N \times N'}$ is the automatically learned global graph content matrix in which each row $\mathbf{C}_{(i, \cdot)} \in \mathbb{\mathbb{R}}^{N'}$ is equivalent to a node of the source graph $G$ and each column $\mathbf{C}_{(\cdot, j)} \in \mathbb{\mathbb{R}}^{N}$ is corresponding to a cluster node of the target coarsened graph $G'$.

The GCont bridges the gaps between the source graph and the target graph and maintains the consistency. On one hand, the elements in $\mathbf{C}$ reflect the interaction between nodes from source graph and clusters from coarsened graph. On the other hand, they contain the graph pattern properties cohered before and after coarsening, thus facilitating generalization across graphs with the same form of features. 

\subsubsection{Attention Assignment using MOA}\label{sec:att-relax}
HAP intends to achieve graph downsampling through a cross-level attention-based aggregator for information interaction between the source graph and the coarsened target graph utilizing global graph property guidance. However, we observe that both HSA and SSA described in Sec.~\ref{sec:attention} only focus on one single graph while CA does not utilize any global information. Although MA introduces master information into the attention process, it is highly affected by the manmade master function. To that end, we propose a new variant of attention mechanism called \emph{Master-Orthogonal-Attention (MOA)}.

\textbf{Computation of Attention Assignment:} 
The input of MOA mechanism is a well-learned representation matrix $\mathbf{H}=\left \{{h}_{1}, {h}_{2}, \cdots, {h}_{N}\right \}$, ${h}_{i} \in \mathbb{R}^{F}$, where $N$ is the number of nodes of the source graph $G$, and $F$ is the feature dimension for each node. Then the graph coarsening module produces a new coarsened graph representation matrix $\mathbf{H}'=\left \{{h}'_{1}, {h}'_{2}, \cdots, {h}_{N'}\right \}$, ${h}' _{i} \in \mathbb{R}^{F}$ as its output, where $N'$ is the number of clusters of the coarsened graph. Each cluster will then be regarded as an individual node. Meanwhile, adjacent matrix $\mathbf{A}\in \mathbb{\mathbb{R}}^{N \times N}$ will also be updated to $\mathbf{A'}\in \mathbb{R}^{{N'} \times {N'}}$. Please note that the number of graph coarsening modules and the coarsened graph size $N'$ are determined by the real application graph data. In our experiment, we employ two coarsening modules and we evaluate it in the experiment.

After having obtained the global graph content matrix, we can employ an orthogonal\footnote{The terminology ``orthogonal'' here means rows and columns of a 2D matrix, which is different from the meaning of orthogonal vectors in a mathematical sense.} cross-level attention mechanism between nodes of the source graph and clusters of the target coarsened graph. The attention matrix $\mathbf{M}\in \mathbb{R}^{
N\times N'}$ is formed with elements as follows:
\begin{equation}\label{eq:OA attention}
    \mathbf{M}_{ij}=\sigma\left(\mathbf{a}^{\top}\left [ \mathbf{C}_{{Row}_{i}}\Vert \mathbf{C}_{{Col}_{j}}\right]\right)
\end{equation}
where $\mathbf{\sigma}$ is the \emph{LeakyReLU} nonlinearity, $[\cdot \Vert \cdot]$ is a concatenation operation with relaxed dimension of $\mathbf{C}_{(\cdot, j)}$ from $\mathbb{R}^N$ to $\mathbb{R}^{N'}$, and $\mathbf{a}^{\top}\in \mathbb{R}^{2N'}$ is the trainable shared attentional parameter with relaxed dimension $2N'$. The reason for the relaxation will be given below.

$\mathbf{M}$, which is equivalent to a cross-level aggregator, offers a fully-connected information channel between the source-graph nodes and the target coarsened-graph clusters, with each element $\mathbf{M}_{ij}$ indicating the importance of node $i$'s feature to cluster $j$. The local substructure is preserved by attention mechanism while the high-order dependency is also captured through the fully-connected information channel, thus strengthening feature reservations. We normalize it for better evaluation:
\begin{equation}\label{eq:normalize}
    \mathbf{M}=\frac{exp \left ( \mathbf{M}_{ij} \right )}{\sum_{\mathbf{k}\in {N'}}exp\left ( \mathbf{M}_{ik} \right)}.
\end{equation}

MOA mechanism synthesizes both self-attention and cross-attention with master-attention. On one hand, the proposed MOA mechanism calculates the attention coefficients based on the GCont alone, so we can sort it into self-attention mechanism. On the other hand, the attention is predicted between the source graph and the target coarsened graph, so we may also classify it as cross-attention mechanism.

\textbf{Relaxation of Attentional Parameter: }
In traditional graph attention scheme~\cite{velivckovic2017graph}, attention coefficients is calculated as follows:

\begin{equation}
	\mathbf{M}_{ij}=\sigma\left(\mathbf{a}^{\top}\left [ \mathbf{W}h_{i}\Vert\mathbf{W}h_{j} \right]\right)
\end{equation}
where $\mathbf{\sigma}$ is the \emph{LeakyReLU} nonlinearity, $\mathbf{a}^{\top}\in \mathbb{R}^{2F'}$ is the trainable shared attentional mechanism, $\mathbf{W}\in \mathbb{R}^{F'\times F}$ is a weight matrix to produce new node features from cardinality $F$ to $F'$, $h_{i}\in \mathbb{R}^{F}$ and $h_{j}\in \mathbb{R}^{F}$ are the input node features, and $[\cdot \Vert \cdot]$ is a concatenation operation. 

Apparently, the trainable shared attentional parameter $\mathbf{a}^{\top}\in \mathbb{R}^{2F'}$ in the conventional graph attention mechanism is irrelevant to the node number of the input graph. However, in our MOA mechanism, the dimension of $\mathbf{C}_{(\cdot, j)} \in \mathbb{R}^{N}$ is related to the node number $N$ of the inputted source graph, making the concatenation $\left[ \mathbf{C}_{(i,\cdot)}\Vert \mathbf{C}_{(\cdot, j)}\right] \in \mathbb{R}^{{N+N'}}$. As a result, the trainable shared attentional mechanism would be initialized as $\mathbf{a} \in \mathbb{R}^{{N+N'}}$, which is sensitive to the node number $N$ of the inputted source graph. 

Manifestly, $N$ varies with the input and it is unknown for parameter initializing stage. Thus, proper relaxation offers intriguingly good performance when standard techniques appear to suffer. We loose $N$ to $N'$, so that $\left[\mathbf{C}_{(i,\cdot)}\Vert \mathbf{C}_{(\cdot, j)}\right]\in \mathbb{R}^{2N'}$. We theoretically analyse the validity of the relaxation on the prediction outcome in Sec.~\ref{sec:analysis}.

\subsubsection{Cluster Formation}
The learning of the global graph content and the cross-level aggregator constitutes a concordant unity, and complement and restrict mutually. Subsequently, we generate the coarsened graph representation matrix $\mathbf{H'}\in \mathbb{R}^{N' \times F}$ and update the adjacent matrix $\mathbf{A}'\in \mathbb{R}^{N' \times N'}$:
\begin{equation}\label{eq:H'}
    \mathbf{H'}=\mathbf{Aggregate}\left(\mathbf{H}\right)=\mathbf{M}^{\top}\mathbf{H},
\end{equation}
\begin{equation}\label{eq:A'}
    \mathbf{A'}=\mathbf{M}^{\top}\mathbf{A}\mathbf{M}.
\end{equation}

\subsubsection{Soft Sampling}
According to Lee et al.~\cite{lee2019self-attention}, handling adjacent data with a sparse matrix in GNN contributes to decreasing the computational complexity from $\mathcal{O}(|\mathcal{V}|^{2})$ to $\mathcal{O}(|\mathcal{E}|)$ and also reduces space complexity. However, the adjacent matrix $\mathbf{A'}$ turns to be a dense one from a sparse assignment. That said, the structure of the source graph $G$ is refined to a fully-connected downsampled one. Proper edge sampling will lead to saving both time and storage without a dramatic loss of accuracy. As a workaround, we adopt the Gumbel-SoftMax~\cite{jang2017categorical} to achieve soft sampling for neighborhood relationship, thus decreasing edge density for the sampled adjacent matrix $\mathbf{\widetilde{A'}}$:
\begin{equation}\label{eq:glumbel}
    \mathbf{\widetilde{A'}_{ij}}=\frac{exp\left(\left(log\mathbf{A}_{ij}+\mathbf{g}_{ij}\right)/\mathbf{\tau}\right)}{\sum_{k=1}^{N'}exp\left(\left(log\mathbf{A}_{ik}+\mathbf{g}_{ik}\right)/\mathbf{\tau}\right)}
\end{equation}
where $\mathbf{g}=-log\left(-log\left(\mathbf{u}\right)\right)$ and $\mathbf{u}\sim \mathbf{Uniform}\left(0,1\right)$. Here, we set the softmax temperature parameter $\mathbf{\tau}=0.1$ to make the adjacent matrix distribution close to one-hot. This operation reduces edge density as much as possible but preserves the connectivity of graphs.

\begin{algorithm}[tb]
\caption{Algorithm of graph coarsening module}
\label{alg:oamodule}
	\leftline{\textbf{Input:} $\mathbf{H}$ - node feature matrix of $G$, }
	\leftline{ \quad\quad \ \ \ $\mathbf{A}$ - adjacent matrix of $G$,}
	\leftline{ \quad\quad \ \ \  $N$ - node number of $G$,}
	\leftline{ \quad\quad \ \ \ $N'$ - node number of $G'$}
	\leftline{\textbf{Output:} $\mathbf{H}'$ - node feature matrix of coarsened graph $G'$, }
	\leftline{ \quad\quad\quad \  $\mathbf{\widetilde{A'}_{ij}}$ - the updated adjacent matrix with soft-sampling}
		
\begin{algorithmic}[1] 
	\State Initialize the global graph content $\mathbf{C}$\Comment{cf. Equation (\ref{eq:global graph content})}
	
	\For{$i=1,\cdots,N$}
	
	\For{$j=1,\cdots,N'$}
	
	\State Generate cross-attention coefficient $\mathbf{M}_{ij}$ for each 
	
	\Statex \quad \quad \quad cross-level node-cluster pair\Comment{cf. Equation (\ref{eq:OA attention})}
	
	\EndFor
	
	\EndFor
	
	\State Define $\mathbf{M}$ with $\mathbf{M}\left(i,j\right)=\mathbf{M}_{ij}, {\forall}i \in \left\{1,2,\cdots,N\right\}, {\forall}j \in \left\{1,2,\cdots,N'\right\}$
	\State Normalize the cross-level attention assignment$\mathbf{M}$ 	\Statex \Comment{cf. Equation (\ref{eq:normalize})}
	
	\State Generate the coarsened graph representation matrix $\mathbf{H}'$ 
	
	\Statex \Comment{cf. Equation (\ref{eq:H'})}
	
	\State Update the adjacent matrix $\mathbf{A}'$ for $G'$ \Comment{cf. Equation (\ref{eq:A'})}
	
	\For{element in $\mathbf{A}'$}
	
	\State Apply Gumbel-Softmax to decrease edge density  
	
	\Statex \Comment{cf. Equation (\ref{eq:glumbel})}
	
	\EndFor
	
	\State \textbf{Return} $\mathbf{H}',\mathbf{\widetilde{A'}_{ij}}$
\end{algorithmic}
\end{algorithm}

\subsection{Learning}
The proposed HAP supports three types of input: single graph $G$ for graph classification, pairwise graphs $\left(G_{1},G_{2}\right)$ for graph matching, and triplet graphs $\left(G_{1},G_{2},G_{3}\right)$ for graph similarity learning. All of the input graphs will be coarsened to a 1D vector at the final graph embedding layer, which can be used to compute graph similarity directly. 
Meanwhile, as is demonstrated in model structure, HAP alternates between node embedding and graph coarsening, thus generating different graph representation matrix $\mathbf{G}_{i}^{k}$ at graph coarsening layer $\mathbf{k}$. As a result, we also propose a hierarchical similarity measure by jointly utilizing hierarchical graph representations.

\subsubsection{Prediction}
For graph classification tasks with a single input graph $G$, the final graph representation $\mathbf{G}$ is directed fed into two fully-connected layers with a $softmax \left(\cdot\right)$ activation on the output to get the predicted label $\hat{Y}$. Then we optimize the model with a standard cross-entropy on the graph that has ground truth labels $Y$. The fully-connected layers and the objective function can be represented separately as follows:
\begin{equation}\label{eq:fc}
\left\{\begin{array}{l}
\mathbf{f}_{1}=\sigma\left(\mathbf{W}_{1}\mathbf{G}+\mathbf{b}_{1}\right)\\
\mathbf{f}_{2}=\sigma\left(\mathbf{W}_{2}\mathbf{f}_{1}+\mathbf{b}_{2}\right)
\end{array}\right.
\end{equation}
\begin{equation}\label{eq:loss-single}
    \mathcal{L}_{single}=-\sum_{g\in\mathcal{B}}\sum_{m=1}^{c}Y_{m}^{g}log\hat{Y}_{m}^{g}
\end{equation}
where $\mathbf{W}_{i}$ and $\mathbf{b}_{i}$ represent weights and biases in the $i$-th fully-connected layer respectively for $i\in\left \{ 1,2 \right \}$. $\sigma$ is the adopted \emph{ReLU} and \emph{Softmax} activation function for $\mathbf{f}_{1}$ and $\mathbf{f}_{2}$ separately. $\mathcal{B}$ is the training set of single graphs and $c$ denotes the number of classes.

For graph matching tasks with pairwise input graphs $\left(G_{1}, G_{2}\right)$, pairs are labeled with $true$ or $false$ representing similar or dissimilar respectively. We optimize the normalization function to push the model to convert graph distances to similarity scores with distribution $s\in\left(0,1\right)$:
\begin{equation}\label{eq:norml-for-loss}
    s_{\left(G_{1},G_{2}\right)}^{k}=exp(-scale\times {d}_{\left(G_{1},G_{2}\right)}^{k})
\end{equation}
where $scale\in\left(-\infty,0\right)$ denotes a softmax parameter sensitive to different range of distances and is determined by the real application graph data. Basically, we set it to 0.5. ${d}_{\left(G_{1},G_{2}\right)}^{k}$ represents graph distances of graph pair $\left(G_{1},G_{2}\right)$ at coarsen level $k\in{K}$, and here we use Euclidean distance. Then the model is optimized by hierarchical cross-entropy function as follows:
\begin{equation}\label{eq:loss-pair}
    \mathcal{L}_{pair}=-\frac{1}{K\left|\mathcal{P}\right|}\sum_{k\in K}\sum_{\left(G_{1},G_{2}\right)\in\mathcal{P}}Y_{p}log\left(s_{\left(G_{1},G_{2}\right)}^{k}\right)
\end{equation}
where $\mathcal{P}$ is the training set of pairwise graphs. $Y_{p}\in\left \{ 0,1 \right \}$ is the label for this pair.

For graph similarity learning tasks with triplet input graphs $\left(G_{1}, G_{2}, G_{3}\right)$, hierarchical Mean Squared Error (MSE) loss function is employed as follows:
\begin{equation}\label{eq:loss-triple}
    \mathcal{L}_{triple}=\frac{1}{K\left|\mathcal{T}\right|}\sum_{k\in K}\sum_{t\in\mathcal{T}}\left(\left({d}_{\left(G_{1},G_{2}\right)}-{d}_{\left(G_{1},G_{3}\right)}\right)-\mathcal{T}_{GED}\right)
\end{equation}
where $\mathcal{T}$ is the training set of triplet graphs, $\mathcal{T}_{GED}$ denotes ground truth triplet proximity defined by relative Graph Edit Distance (GED) at Sec.~\ref{sec:triplet}. 

\subsubsection{Hierarchical Prediction}
As shown in \figurename~\ref{fig:model structure}, we adopt a hierarchical prediction strategy to further facilitate the training process and fully utilize the hierarchical intermediate features of coarsened  graphs. The outputs of every coarsening process are summarized as the intermediate graph feature, which will be fed into the learning module for graph matching or graph similarity learning.

\section{Theoretical Analysis}

\subsection{Computational Complexity Analysis}
In the following, we theoretically analysis the computational complexity of the proposed HAP and show the superiority of the proposed graph coarsening module.

\begin{myclaim}[Time Complexity]
	The time complexity of the proposed HAP with $K$ graph coarsening modules in dowmsampling ratio $r$ is approximately $\mathcal{O}(N^{2})$, where $N$ is the number of nodes of the original input graph.
\end{myclaim}

\begin{myPro}
	The time complexity of HAP involves three parts corresponding to the three stages of GNN-based graph-level representation learning models: (1) node embedding; (2) graph coarsening; and (3) learning. The time complexity of node embedding stage is $\mathcal{O}(NFF'+\mathcal{|E|}F')$~\cite{velivckovic2017graph}, where $F'$ is the dimension of output node features. After that, to downsample node number in the $k$-th graph coarsening module, where $k \in \left\{1,2,\cdots, K\right\}$, it requires $\mathcal{O}(r^{k-1}N\cdot r^{k}N)$. Let's suppose $r$ remains constant among all the coarsening modules. Then the time complexity for all the $K$ graph coarsening modules is $\mathcal{O}(rN^{2}+r^{3}N^{2}+\cdots+r^{2K-1}N^{2})$. Due to the fact that $r$ is less than 1, $r^{3}+\cdots+r^{2K-1}$ is a couple of orders of magnitude smaller than $r$. So the time complexity of graph coarsening stage is roughly equivalent to $\mathcal{O}(rN^{2})$. Eventually, for the learning stage, the time complexity is $\mathcal{O}(F_{G}^{2})$, where $F_{G}$ is the dimension of the graph level embedding for the input graph $G$. Therefore, the overall computational complexity of the proposed HAP framework is $\mathcal{O}(rN^{2}+NFF'+\mathcal{|E|}F'+F_{G}^{2}) \approx \mathcal{O}(rN^{2}) \approx \mathcal{O}(N^{2})$.
\end{myPro}

Specifically, when a proper coarsening ratio $r$ is chosen where $rN \ll N$ (e.g., $r=0.05$ and $N=100$), the actual execution time of the proposed HAP will become almost linear to $N$.


\subsection{Permutation Invariance}
Graph pooling methods need to be permutation invariant since they should guarantee that the graph-level representation does not vary with the input order of node-level representations. As for the proposed graph coarsening module, we proof that it is graph permutation invariant.

\begin{myDef}[Permutation matrix]
	$\mathbf{P}_{n} \in \left\{0,1\right\}^{n \times n}$	 is a permutation matrix of size $n$ iff $\sum_{i}{\mathbf{P}_{i,j}}=1$ $ \forall j$ and $\sum_{j}{\mathbf{P}_{i,j}}=1$ $ \forall i$.
\end{myDef}

\begin{myclaim}[Permutation invariance]
	Let $\mathbf{P}_{n} \in \left\{0,1\right\}^{n \times n}$ be any permutation matrix, $G=\left(\mathbf{A}, \mathbf{X}\right)$ be any undirected graph, a function $f\left(\mathbf{A},\mathbf{X}\right)$ be a pooling operation depending on graph $G$, graph permutation is defined as $f\left(\mathbf{A},\mathbf{X}\right)=f\left(\mathbf{P}_{n}\mathbf{A}\mathbf{P}_{n}^{\top}, \mathbf{P}_{n}\mathbf{X}\right)$. The proposed graph coarsening module is graph permutation invariant.
	
\end{myclaim}

\begin{myPro}
	$\mathbf{M}$ is computed by an attention mechanism between source nodes and target coarsened clusters. Since the attention function are operated between node set and cluster set, the order of nodes or clusters has no effect to the result, we have:
	\begin{equation}
		\mathbf{M} \rightarrow \mathbf{P}_{n}\mathbf{M}\mathbf{P}_{n}^{\top}
	\end{equation}
	Since ${\mathbf{A}}'=\mathbf{M}^{\top}\mathbf{A}\mathbf{M}$ and any permutation matrix is orthogonal, applying $\mathbf{P}_{n}^{\top}\mathbf{P}_{n}=\mathbf{I}$ to it, we get:
	\begin{equation}
		\mathbf{A}' \rightarrow \mathbf{P}_{n}\mathbf{A}'\mathbf{P}_{n}^{\top}	
	\end{equation}
	Since $\mathbf{X}'=\mathbf{E}^{\top}\mathbf{X}$, applying $\mathbf{P}_{n}^{\top}\mathbf{P}_{n}=\mathbf{I}$ to it, we get:
	\begin{equation}
		\mathbf{X}' \rightarrow \mathbf{P}_{n}\mathbf{X}'
	\end{equation}
	As a result, $f\left(\mathbf{A},\mathbf{X}\right) \rightarrow f\left(\mathbf{P}\mathbf{A}\mathbf{P}^{\top}, \mathbf{P}\mathbf{X}\right)$, HAP is graph invariant.

\end{myPro}


\subsection{Validity of Relaxation for Attentional Parameter}\label{sec:analysis}
In Sec.~\ref{sec:att-relax}, we conduct a relaxation operation $\psi: \mathbf{a} \in \mathbb{R}^{N + N'} \rightarrow \mathbf{a} \in \mathbb{R}^{2N'}$ for the attentional parameter. Substantially, the relaxation is applied to the column dimension of GCont $\mathbf{C}$ during concatenation. A natural question is that whether the relaxation affects the accuracy of attention coefficients, which may directly lead to neglecting important information during cross-level aggregation. We now theoretically analyze this question.

\begin{myDef}[LeakyReLU]
    LeakyReLU is a monotone increasing activation function:
    \begin{equation}
        \varphi (x)=\left\{\begin{matrix}
x & x\geq 0 \\ 
\frac{x}{a} &  x< 0,
\end{matrix}\right.
where~a \in (1, + \infty)
    \end{equation}
\end{myDef}

\begin{myclaim}
	Let $N$ \textgreater $N'$, $\mathbf{C}_{i} \in \mathbb{R}^{N'}$, $\mathbf{C}_{j} \in \mathbb{R}^{N}$ and $\mathbf{a} \in \mathbb{R}^{N + N'}$ be vectors before relaxation, let $\mathbf{C}'_{j} \in \mathbb{R}^{N'}$ and $\mathbf{a}' \in \mathbb{R}^{2N'}$ be vectors after relaxation, let $[\cdot \Vert \cdot]$ be concatenation operation, and \emph{LeakyReLU} be a nonlinearity. Then $\emph{LeakyReLU}\left(\mathbf{a}^{\top}[\mathbf{C}_{i} \Vert \mathbf{C}_{j}]\right) = \emph{LeakyReLU}\left(\mathbf{a}'^{\top}[\mathbf{C}_{i} \Vert \mathbf{C}_{j}']\right)$.

\end{myclaim}

\begin{myPro}
	The essence of $\mathbf{a}^{\top}[\mathbf{C}_{i} \Vert \mathbf{C}_{j}]$ is a similarity comparison between vector $\mathbf{C}_{i} \in \mathbb{R}^{N'}$ and vector $\mathbf{C}_{j} \in \mathbb{R}^{N}$. Due to the reason that vectors with different dimensions are non-comparable, the lacking dimension needs to be padded with zero. So that:
	\begin{equation}
		\mathbf{C}_{i}=\left(c_{1}, \cdots, c_{N'}\right) \rightarrow \mathbf{C}'_{i}=\left(c_{1}, \cdots, c_{N'},  \underbrace{0, 0, \cdots, 0}_{N - N'}\right)
	\end{equation}
	While do comparison between 	$\mathbf{C}_{i}$ and $\mathbf{C}'_{j}$, we can also pad them as follows:
	\begin{equation}
		\mathbf{C}_{i}=\left(c_{1}, \cdots, c_{N'}\right) \rightarrow \mathbf{C}'_{i}=\left(c_{1}, \cdots, c_{N'},  \underbrace{0, 0, \cdots, 0}_{N - N'}\right)
	\end{equation}
	\begin{equation}
		\mathbf{C}'_{j}=\left(c_{1}, \cdots, c_{N'}\right) \rightarrow \hat{\mathbf{C}}_{j}=\left(c_{1}, \cdots, c_{N'},  \underbrace{0, 0, \cdots, 0}_{N - N'}\right)
	\end{equation}
	
	Hence, $\mathbf{a}^{\top}[\mathbf{C}_{i} \Vert \mathbf{C}_{j}]=\mathbf{a}^{\top}[\mathbf{C}_{i} \Vert \mathbf{C}'_{j}]$. Based on known conditions that $\emph{LeakyReLU}$ is monotonically increasing, so that $\emph{LeakyReLU}\left(\mathbf{a}^{\top}[\mathbf{C}_{i} \Vert \mathbf{C}_{j}]\right) = \emph{LeakyReLU}\left(\mathbf{a}'^{\top}[\mathbf{C}_{i} \Vert \mathbf{C}_{j}']\right)$.
	
	
\end{myPro}

As a result, the relaxation for attentional parameter has no negative effects for the attention computation and feature extraction.

\section{Experiments and Evaluation}\label{sec:experiment}
We evaluate HAP against a number of state-of-the-art methods to answer the following questions:

\textbf{Q1:} How does HAP compare with other baselines when evaluated with downstream tasks including graph classification, graph matching and graph similarity learning? (Sec.~\ref{sec:task1}, Sec.~\ref{sec:task2}, Sec. ~\ref{sec:task3})

\textbf{Q2:} How dose the original HAP compare with ablated ones with graph coarsening module replaced by other state-of-the-art pooling algorithms? (Sec.~\ref{sec:pooling-ablation})

\textbf{Q3:} How does the number of the graph coarsening modules influence the quality of graph-level representations generated by HAP? (Sec.~\ref{sec:effect-of-moas})

\textbf{Q4:} Do key designs of HAP contribute to better generalization performance? (Sec.~\ref{sec:genera})


\begin{table}[h]
\setlength{\abovecaptionskip}{0.cm}
\setlength{\belowcaptionskip}{-0.cm}
\setlength\tabcolsep{1.5pt}
  \centering
  \caption{Statistics of datasets. IMDB-B, IMDB-M, COLLAB, MUTAG, PROTEINS and PTC are for graph classification; AIDS and LINUX are for graph similarity learning; and the Synthetic data is for graph classification.}
  \scalebox{1}{
    \begin{tabular}{lcccccc}
    \toprule
   \textbf{Dataset} &\textbf{\#Graphs} &\textbf{\#Triples} &\textbf{\#Pairs} &\textbf{Max.$\mathcal{V}$} &\textbf{Avg.$\mathcal{V}$} &\textbf{\#Classes}\\
    \midrule
    IMDB-B &1000 &- &- &136 &19.8 &2\\
    IMDB-M &1500 &- &- &89 &13.0 &3\\
    COLLAB &5000 &- &- &492 &74 &3\\
    MUTAG &188 &- &- &28 &17.9 &2\\
    PROTEINS &1113 &- &- &620 &39.1 &2\\
    PTC &344 &- &- &109 &25.5 &2\\
    \hline
    AIDS &- &171900 &- &10 &8.9 &-\\
    LINUX &- &409600 &- &10 &7.7 &-\\
    \hline
    Synthetic data &- &- &8750 &300 &105.7 &2\\
    \bottomrule
    \end{tabular}%
    }
  \label{tab:data}%
\end{table}%

\subsection{Experimental Setup}
\subsubsection{Datasets}
We perform experiments on eight real-world datasets and one synthetic dataset varying with tasks. The graph statistics are summarized in \tablename~\ref{tab:data}. 

Evaluating graph matching task requires benchmark datasets with ground-truth labels (true for matching and false for unmatching). To the best of our knowledge, no public-available real-world dataset holds such ground-truth labels. To fill this gap, we conduct a synthetic dataset, a collection of labeled graph pair $\left(G_{1},G_{2}\right)$ with edge probability $p\in \left[0.2, 0.5\right]$ generated by the VF2 graph matching library~\cite{cordella2004sub}. Given a graph $G$, a positive sample is the maximum connected subgraph randomly extracted with 1 to 3 nodes less than $G$. And a negative sample is created by randomly adding 3 to 7 nodes with the same edge probability.

\subsubsection{Baselines}
We compare HAP with three kinds of baselines:

\textbf{Graph pooling baseline:} For comparison of total pooling, we choose GCN-concat (concatenation of GCN-based node-level representations), SumPool~\cite{xu2019how}, MeanPool, MeanAttPool~\cite{bai2019simgnn}, and Set2Set~\cite{vinyals2015order}. For \emph{TopK} pooling, we use SortPooling~\cite{zhang2018end}, AttPool~\cite{huang2019attpool}, gPool~\cite{gao2019graph} and SAGPool~\cite{lee2019self}. For group pooling, we compare with DiffPool~\cite{Ying2018Hierarchical} and ASAP~\cite{ranjan2020asap}. We also conduct evaluation on an unsupervised method StructPool~\cite{yuan2019structpool}.

\textbf{Graph matching baseline:} We focus on the Graph Matching Network (GMN)~\cite{li2019graph} specifically designed for pairwise graph similarity learning.

\textbf{Graph similarity learning baseline:} There are two categories of graph similarity learning baselines. Due to the reason that the ground-truth triplet proximity for graph similarity learning task is calculated by conventional rigorous GED algorithm, the first type is referred to as conventional approximate GED algorithms for comparison, including Beam search~\cite{neuhaus2006fast}, VJ~\cite{fankhauser2011speeding} and Hungarian~\cite{riesen2009approximate} algorithm. The other type includes SimGNN~\cite{bai2019simgnn} and GMN, which are GNN-based models.

\subsubsection{Parameter Settings}
For the basic model structure of HAP, we set two node \& cluster embedding layers before every following graph coarsening module, and a total of two coarsening modules are needed. 
Adma optimizer is used with initial learning rate 0.01 for graph classification datasets, 0.0015 for AIDS, 0.0001 for LINUX and synthetic data. For social network datasets IMDB and COLLAB with no informative node features, we use one-hot encoding of node degrees as initial node input. Similarly, we adopt one-hot encoding of node labels for AIDS dataset, while others are initialized identically. For graph classification and other tasks, the initial dimension is 64 and 128, respectively. All of the datasets are randomly partitioned into 8:1:1 for training/validation/testing. For all the baseline methods, we conduct experiments under the default settings reported in the original work.

\begin{table*}[t]
\centering
\setlength\tabcolsep{7pt}
\caption{Graph classification accuracy in percent.}
\label{tab:classification}
\begin{tabular}{c|l|cccccc}
\toprule
\textbf{Method}    & {\textbf{Model}} & { \textbf{IMDB-B}} & { \textbf{IMDB-M}} & { \textbf{COLLAB}} & { \textbf{MUTAG}} & { \textbf{PROTEINS}} & { \textbf{PTC}} \\
\midrule
                                          & GCN-concat                                                & 74.01                                   & 48.03                                   & 63.22                                  & 72.22                                & 70.27                                   & 58.82                              \\
                                          & SumPool                                                   & 76.02                                   & 52.01                                   & 72.83                                  & 89.47                                & 73.21                                   & 68.57                              \\
                                          & MeanPool                                                  & 74.02                                   & 51.33                                 & 71.26                                 & 85.01                                  & 72.32                                   & 63.89                              \\
                                          
 & MeanAttPool                                               & 75.01                                   & 52.03                                   & 72.65                                 & 85.06                                  & 73.21                                   & 63.89                              \\
\multirow{-5}{*}{\textbf{Universal pooling}}                                          & Set2Set                                                   & 68.02                                   & 50.66                                 & 64.23                                  & 88.89                                & 71.17                                   & 55.88                              \\
\hline
                                          & SortPooling                                               & 66.83                                  & 47.02                                   & 72.94                                  & 83.33                                & 74.05                                   & 56.47                              \\
                                          & AttPool-global                                            & 70.13                                  & 47.53                                 & 77.36                                 & 86.67                                & 74.68                                   & 66.18                              \\
                                          & AttPool-local                                             & 70.83                                  & 48.73                                 & 79.12                                  & 82.22                                & 74.77                                   & 66.47                              \\
                                          & gPool                                                     & 78.02                                   & 54.67                                 & \textbf{82.25}          & 87.72                                & 73.87                                   & 68.57                              \\
                                         
\multirow{-5}{*}{\textbf{TopK   pooling}} & SAGPool                                                   & 75.03                                   & 48.64                                  & 78.22                                  & 75.02                                  & 72.97                                   & 47.06                              \\

\hline
                                          & DiffPool                                                  & 77.04                                   & 52.03                                   & 61.87                                 & 77.78                                & 73.87                                   & 58.82                              \\
                                          
\multirow{-2}{*}{\textbf{Group pooling}}  & ASAP                                                      & 73.04                                   & 50.13                                  & 80.52                                  & 78.35                                 & 73.92                                    & 58.01                                \\
\hline
\textbf{Unsupervised}                     & StructPool                                                & 74.06                                   & 53.33                                 & 70.85                                  & 83.33                                & 72.07                                   & 67.64                              \\
\hline
                                          & \textbf{HAP   (ours)}                                              & \textbf{79.04}                                   & \textbf{55.33}                                 & 73.95                                  & \textbf{95.01}                                  & \textbf{76.46}                                   & \textbf{69.44} \\
                                                                    
\bottomrule
\end{tabular}
\end{table*}

\subsection{Task1: Graph Classification}\label{sec:task1}
We evaluate HAP on six benchmark graph classification datasets and compare it with several state-of-the-art approaches belonging to different pooling categories respectively. For AttPool, we try different attention functions (global attention and local attention) to obtain the graph-level representations. For HAP, we try GAT and GCN for node \& cluster embedding operation and report the better accuracy. Table~\ref{tab:classification} shows the accuracy with the best results marked in bold. We can observe that HAP obtains the best performance on five out of six datasets with an average improvement of 5.9\%.

Of all the graph pooling methods, universal pooling approaches
are the most straightforward ones but achieve considerable effect, especially the SumPool which is consistent in underlying concept with our HAP. Intuitively, higher the quality of graph-level representations, better the graph classification result. The element-wise sum aggregator in SumPool tends to capture all node features in consideration of higher-order node dependency, but the generated graph-level representations fail to obtain sufficient quality, i.e., the quality of graph-level representations is not positively associated with how much node features are acquired. Irrelevant features that may interfere the results are obtained without reducing the weights, thus the final graph-level representations mixed with excessive irrelevant information is detrimental to the graph classification accuracy.

\emph{TopK} pooling approaches produce score-based representations that drop nodes from the original graph with lower scores. As a result, potentially valuable information attached with these nodes and the related substructures may be ignored. From \tablename~\ref{tab:classification}, the performance of \emph{TopK} pooling approaches are universally inferior than other methods that capture more features or structural information. More damaging, SortPool and AttPool-global fail to return a result within 72 hours in practical execution. But there is an exception to the rule: gPool, with consistently better performance than other methods, even excels HAP on COLLAB. gPool computes scores by the multiplication of node feature matrix and a trainable projection vector, so that feature of each node is covered in the estimated scalar projection values by assigning with different weights. This crucial ingredient leads to the outstanding performance. As for the incredible performance on COLLAB, it might be due to the nature of COLLAB dataset. COLLAB covers scientific collaboration between authors. Nodes represent authors and edges indicate co-author relationship between authors. The classification task of each graph is to estimate the field the corresponding researcher belongs to. In this situation, it can be distinguished easily by the authors with Top-K quantity of papers that may be domain experts, while other unknown authors are actually noisy information. Nevertheless, our HAP still has advantages except for such exceptional circumstances.

For the dataset MUTAG, we observe that our HAP incredibly outperforms all the baselines for an average 12.3\% improvement. The significant test result shows that HAP fits the property of MUTAG adequately. MUTAG is a two-class nitro compound dataset with nodes and edges on behalf of atoms and chemical bonds, respectively. Note that molecules of both classes have the common substructure nitro, so that higher-order information beyond the substructure is the crucial for differentiation, which is correctly handled by HAP.

\begin{figure}[t]
\centering
\subfloat[PROTEINS]{\includegraphics[width=0.5\textwidth]{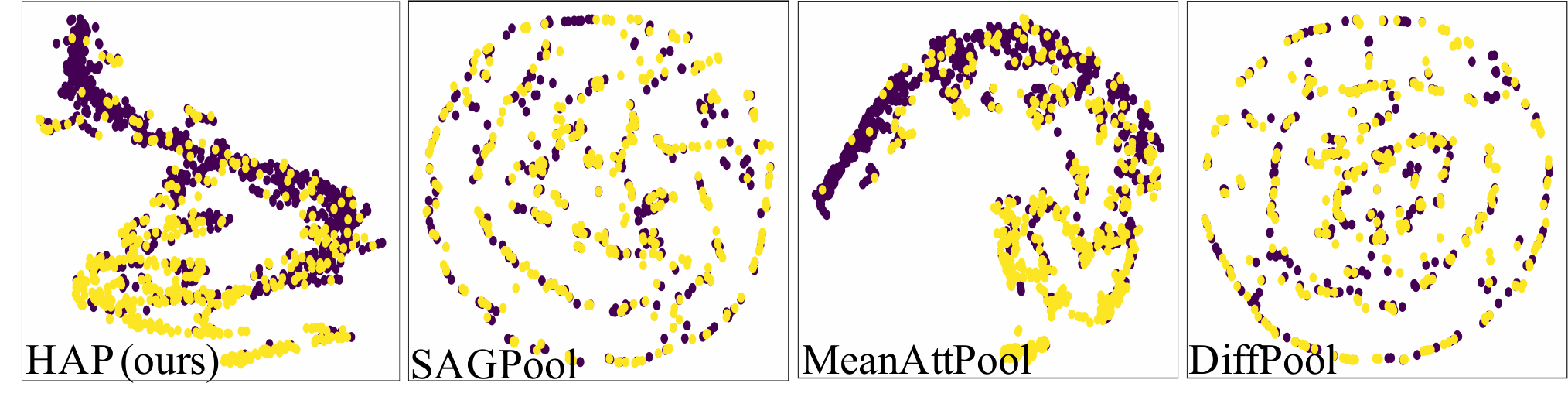}}\\
\vspace{-3mm}
\subfloat[COLLAB]{\includegraphics[width=0.5\textwidth]{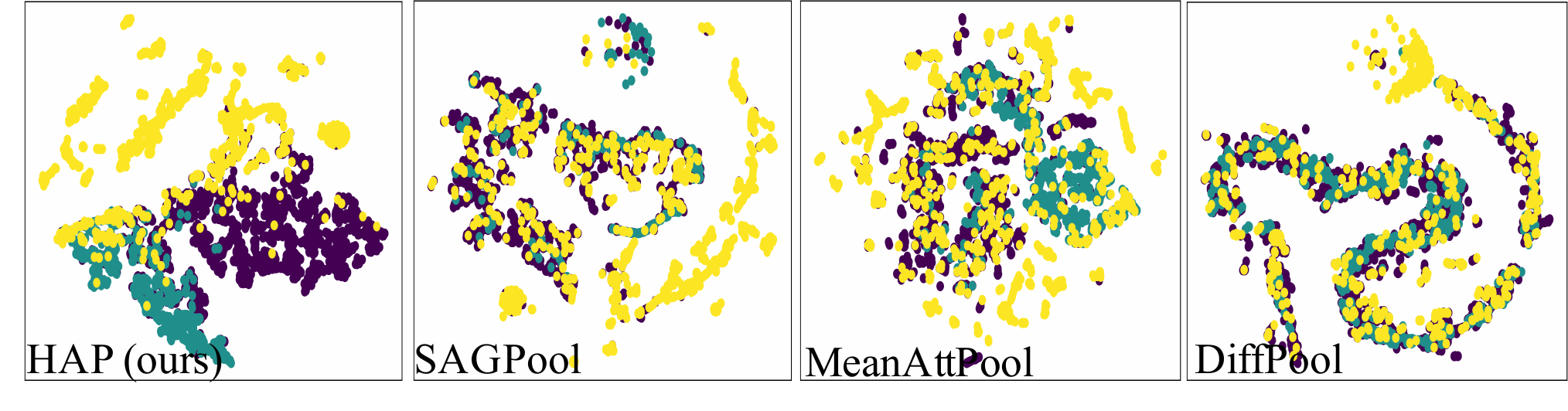}}
\caption{The t-SNE visualization of graph-level representations from HAP and three baselines on PROTEINS (above) and COLLAB (below). Different colors indicate graph samples with different labels.}
\label{fig:vis-methods}
\end{figure}

\textbf{Visualization:} To further conceptualize the effectiveness of the learned graph-level representations, we provide a visualization of the t-SNE on PROTEINS and COLLAB dataset with features extracted by the methods HAP, SAGPool, MeanAttPool and DiffPool (\figurename~\ref{fig:vis-methods}). In each figure, points of different colors exhibits discernible graph clusters with different labels in the projected two-dimensional space. Note that the separability of the cluster border verifies the discriminative power. We can find that HAP performing consistently with MeanAttPool on PROTEINS shows better discriminability of the two classes than SAGPool and DiffPool. As for COLLAB, HAP is far superior to its competitors where three classes are clearly separated, all of which are in accordance with the results suggested in \tablename~\ref{tab:classification}.

\subsection{Task2: Graph Matching}\label{sec:task2}
Four synthetic datasets are generated with different data size $|\mathcal{V}| \in \left\{20, 30, 40, 50 \right\}$ for graph matching task. \tablename \ref{tab:graph-matching} shows the graph matching results w.r.t. graph size.

GMN, specifically designed for graph matching task, makes the node embedding phase dependent on the pair through a cross-graph attention mechanism. However, as shown in \tablename \ref{tab:graph-matching}, HAP drastically boosts the matching accuracy up to 3.5\% compared to GMN on graph size $|\mathcal{V}|=20$. When increasing graph size, HAP achieves a steady raising while GMN decreases gracefully from graph size $|\mathcal{V}|=30$ to $|\mathcal{V}|=40$. This shows the key point: basic node embedding models have been perfectly capable of getting high-quality node-level representations. On the contrary, the core to enhancing graph matching accuracy is to improving the quality of graph-level representations. After replacing the basic pooling module in GMN, the performance of GMN-HAP grows tremendously to be comparable with HAP, further confirming the strong ability of the proposed graph coarsening module.



\begin{figure}[t]
\centering
\includegraphics[width=0.45\textwidth]{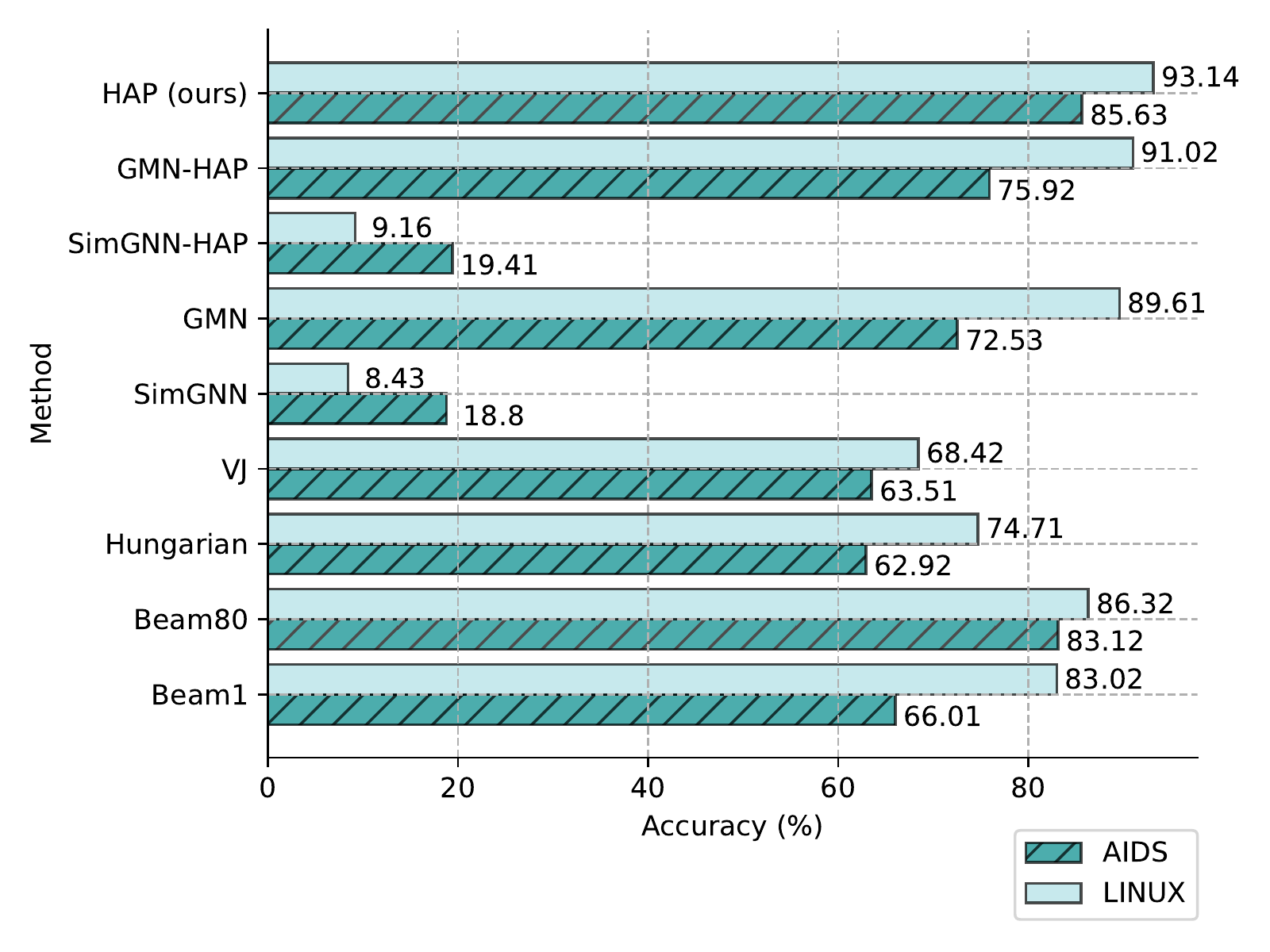}
\caption{Graph similarity accuracy in percent. The triplet similarity based on Beam1, Beam80, Hungarian and VJ is reflected by whether the relative GED is positive or negative.}
\label{fig:graph-similarity}
\end{figure}

\begin{table}[t]
\setlength\tabcolsep{10pt}
\centering
\caption{Graph matching accuracy in percent varying with graph size. GMN-HAP is a variant of GMN where the pooling algorithm in GMN is replaced by the graph coarsening module of HAP.}
\label{tab:graph-matching}
\begin{tabular}{lcccc}
\toprule
Model & $|\mathcal{V}|$=20 & $|\mathcal{V}|$=30 & $|\mathcal{V}|$=40 & $|\mathcal{V}|$=50  \\
\midrule
GMN                       & 95.01 & 97.82 & 96.84 &99.41    \\
GMN-HAP                   & 98.22 & 98.83 & 98.41 &99.82    \\
\textbf{HAP (ours)}                       & \textbf{98.42}& \textbf{98.83} & \textbf{99.43} & \textbf{99.83}  \\ 
\bottomrule
\end{tabular}	
\end{table}


\subsection{Task3: Graph Similarity Learning}\label{sec:task3}
We show the results of HAP for graph similarity learning compared with both conventional approximate GED algorithms and GNN-based models on dataset AIDS and LINUX in \figurename~\ref{fig:graph-similarity}. Note that evaluating graph similarity learning task requires benchmark datasets with ground-truth GEDs processed by the exact algorithm A*. A recent research~\cite{blumenthal2020exact} shows that ``no currently available algorithm manages to reliably compute GED within reasonable time between graphs with more than 16 nodes''. And the experiments on A* show that 10 nodes seem to be reaching the limit of its ability to deal with the problem. To address the gap, we only accept benchmark datasets with the max number of nodes no more than 10 in each graph. Our results demonstrate that HAP is capable of boosting the accuracy 
of state-of-the-art methods.

More specifically, for conventional approximate GED algorithms with high computational complexity, HAP improves accuracy by a relative gain of 16.7\% and 15.1\% on AIDS and LINUX, respectively. In regard to comparing with GNN-based models, HAP is overwhelming to SimGNN, which focuses more on optimizing the exact similarity score between graphs while neglecting the relativity. The result, in one aspect, reflects that a single-minded pursuit of the optimization of pairwise absolute similarity is not necessarily favorable to the relative similarity tasks which are more common in real-world applications to some extent. Similarly, HAP outperforms GMN by a margin of 13.1\% and 3.6\% on AIDS and LINUX, respectively. When replacing pooling methods in SimGNN and GMN with the proposed graph coarsening module, both of them achieve slightly promotion and GMN-HAP obtains comparable accuracy with HAP. These results indicates that our HAP and coarsening module are conducive to a high-quality graph-level representation.

\begin{table*}[t]
\setlength{\abovecaptionskip}{0.cm}
\setlength{\belowcaptionskip}{-0.cm}
\setlength\tabcolsep{1.5pt}
  \centering
  \caption{Ablation study results on graph classification, graph matching and graph similarity learning. HAP-x is a variant of HAP where the graph coarsening module is replaced by x.}
    \begin{tabular}{l|cccccc|cccc|m{1.5cm}<{\centering}m{1.5cm}<{\centering}}
\toprule
\multirow{2}{*}{Ablated Model} & \multicolumn{6}{c|}{Graph Classification}            & \multicolumn{4}{c|}{Graph Matching}                                               & \multicolumn{2}{c}{Graph Similarity Learning} \\ \cline{2-13} 
\rule{0pt}{10pt}                       & IMDB-B & IMDB-M & COLLAB & MUTAG & PROTEINS & PTC   & $|\mathcal{V}|$=20 & $|\mathcal{V}|$=30 & $|\mathcal{V}|$=40 & $|\mathcal{V}|$=50 & AIDS                  & LINUX                 \\ 
\midrule
HAP-MeanPool          & 74.02  & 51.14  & 71.13  & 85.02 & 72.31    & 63.94 & 55.81              & 56.82              & 56.33              & 58.12             & 68.62                 & 66.34                 \\
HAP-MeanAttPool       & 75.03  & 52.25  & 72.64  & 85.13 & 73.22    & 63.91 & 85.23              & 84.01              & 80.63              & 86.62             & 78.31                 & 89.02                 \\
HAP-SAGPool           & 70.34  & 44.21  & 72.15  & 75.22 & 59.86    & 61.13 & 61.55              & 63.62              & 60.13              & 60.05             & 74.42                 & 66.53                 \\
HAP-DiffPool          & 77.35  & 47.12  & 61.92  & 80.03 & 66.12    & 55.67 & 65.93              & 63.04              & 63.82              & 61.95             & 80.73                 & 91.30                 \\
\textbf{HAP (ours)}            &\textbf{79.04} & \textbf{55.33} & \textbf{73.95} & \textbf{95.01} & \textbf{76.46} & \textbf{69.44} & \textbf{98.42}     & \textbf{98.81}     & \textbf{99.43}     & \textbf{99.84}    & \textbf{85.62}        & \textbf{93.11}                 \\
\bottomrule
\end{tabular}
  \label{tab:ablations}%
\end{table*}%

\subsection{Ablation Studies}
\subsubsection{Comparison of Graph Pooling Mechanisms}\label{sec:pooling-ablation}
To study the effectiveness of our proposed graph coarsening module, we fix other components of HAP framework, and replace our coarsening module with other four differentiable graph pooling methods, i.e., MeanPool, MeanAttPool, SAGPool and DiffPool, referring these variants as HAP-MeanPool, HAP-MeanAttPool, HAP-SAGPool and HAP-DiffPool, respectively. The performance of HAP and its four ablated variants on graph classification, graph matching and graph similarity learning task is shown in \tablename~\ref{tab:ablations}.


We observe that compared with other four ablated variants whose performance fluctuates wildly among tasks, our HAP achieves superior performances on all of the twelve datasets for the three tasks. We also find that HAP-MeanPool ranks the bottom across tasks, especially inferior on graph matching and graph similarity learning by a margin of 17\% to 42.61\%. This validates that the multiformity features which may be redundant information in single-input graph classification task is crucial for multiple-input graph-level tasks to do horizontal comparison. On the contrary, HAP-MeanAttPool brings about performance benefits against other ablated variants. This indicates that global-wise information aggregation can be helpful for graph-level representation learning. Further, with the help of the proposed graph coarsening module, our HAP achieves adaptive graph structure sensibility based on a global-wise information aggregation, which utilizes both local structure and global pattern properties, thus contributing to a high-quality graph-level representation. Similarly, when comparing with HAP-DiffPool, a one-hop neighborhood aggregator, HAP can also improve graph-level representation quality by joining high-order dependency among nodes that may hold significant information. Moreover, the performance comparison between HAP-SAGPool and HAP reveals that our HAP can indeed retain key graph information that may be attached to the abandoned nodes.


\begin{table}[h]
\setlength\tabcolsep{2pt}
  \centering
  \caption{The effects of different number of graph coarsening modules in percent.}
\begin{tabular}{l|cccc|m{1.5cm}<{\centering}m{1.5cm}<{\centering}}
\toprule
\multirow{2}{*}{Model} & \multicolumn{4}{c|}{Graph Matching}                                                                        & \multicolumn{2}{c}{Graph Similarity Learning}        \\
\cline{2-7}
        \rule{0pt}{10pt}               & $|\mathcal{V}|$=20 & $|\mathcal{V}|$=30 & $|\mathcal{V}|$=40 & $|\mathcal{V}|$=50 & AIDS & LINUX \\
\midrule  
baseline               & 85.21                    & 84.02                    & 80.65                    & 86.63                    & 75.52                    & 67.63                     \\
Coarsen=1                   & 99.04                    & 97.16                    & 97.61                    & 96.08                    & 83.22                    & 84.83                     \\
Coarsen=2                   & \textbf{99.72}                    & 98.76                    & \textbf{99.78}                    & 98.4                     & 84.11                    & 89.42                     \\
Coarsen=3                   & 97.62                    & \textbf{99.45}                    & 99.35                    & \textbf{99.39}                    & \textbf{85.04}                    & \textbf{90.65}      \\
\bottomrule              
\end{tabular}
\label{tab:MOA module}
\end{table}

\subsubsection{Comparison of Different Number of Graph Coarsening Module}\label{sec:effect-of-moas}
\tablename~\ref{tab:MOA module} shows the performance of graph matching and graph similarity learning by adopting different number of graph coarsening modules in HAP. All experiments are conducted using HAP-MeanAttPool as the baseline with fixed coarsen ratio for the same dataset. We observe that replacing the MeanAttPool with one our graph coarsening module, denoted as Coarsen = 1, improves the performance by at least 10.2\%, which can effectively demonstrate the significance of the our proposed coarsening module. Furthermore, increasing coarsening modules from one to two can improve the performance by at most 5.4\%. Finally, increasing coarsening modules from two to three slightly improves the performance by an average of 0.7\%. These results demonstrate that the proposed graph coarsening module can dramatically improve the performance by coarsening graphs in a hierarchical manner. 

\begin{figure}[h]
\centering
\subfloat[PROTEINS]{\includegraphics[width=0.5\textwidth]{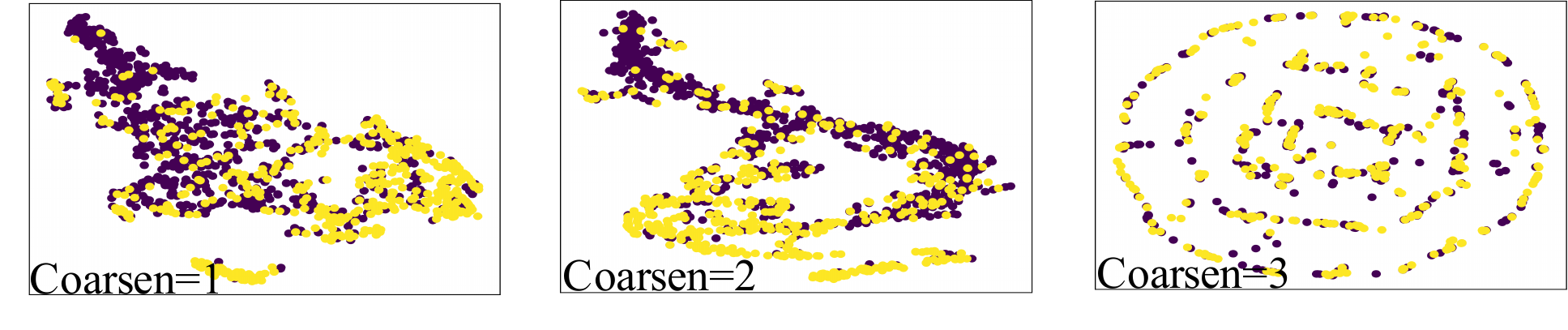}}\\
\vspace{-3mm}
\subfloat[COLLAB]{\includegraphics[width=0.5\textwidth]{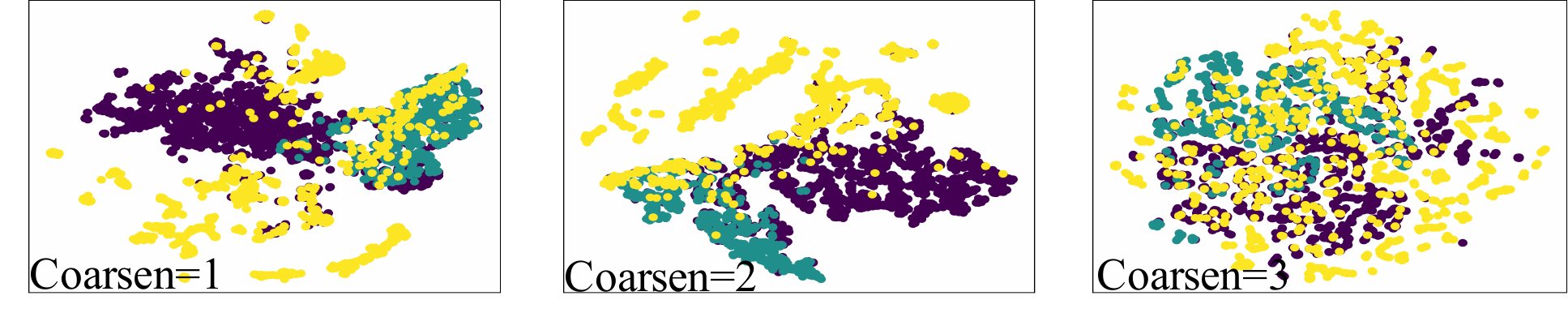}}
\caption{The t-SNE visualization of graph-level representations of HAP with different number of graph coarsening modules on PROTEINS (above) and COLLAB (below). Different colors indicate graph samples with different labels.}
\label{fig:vis-MOAs}
\end{figure}

\textbf{Visualization:} \figurename~\ref{fig:vis-MOAs} visualizes how graph-level representations react with different number of graph coarsening modules in graph classification task. It can be seen that the challenging classification is progressively corrected with the number of graph coarsening modules increasing from one to two, but is easily to be misclassified when there are three coarsening modules.

Synthesizing the above results, the greater the number of graph coarsening modules, the more attached parameters and additional memory usage. To balance the performance and resource usage, we choose Coarsen = 2 as default settings.


\subsubsection{Comparison of Generalization Performance}\label{sec:genera}

While most GNNs are designed to consider the generalization ability to unseen nodes, there are few researches in graph pooling area to address the generalization to unseen graphs. However, in practical applications such as protein molecular structure recognition, researchers are often interested in generalizing the knowledge learned from small-sized molecules to large-sized molecules with the same form of structures. 

In this subsection, we justify the generalization capability of the models by training on small-size graphs and testing on large-sized graphs with the same edge probability for graph matching task. The results shown in \tablename~\ref{tab:generalization} indicate that only HAP can achieve a natural generalization of the small-sized results to the scenarios of large-sized graphs. This is credited to the key strength of HAP: it can effectively learn the global graph content that involves high-level pattern information for the training graph by GCont, thus preserving the pattern properties that are inherited between the training and testing graphs. When applying our graph coarsening module to GMN, GMN-HAP achieves significant improvement of the prediction performance by 8.22\% and 10.31\%, respectively.

\begin{table}[h]
\setlength\tabcolsep{15pt}
\centering
\caption{Generalization performance in percent on graph matching task. Models are trained on graphs with $20 \leq |\mathcal{V}| \leq 50$ and tested on graphs with $|\mathcal{V}|=100$ or $|\mathcal{V}|=200$.}
\label{tab:generalization}
\begin{tabular}{lcc}
\toprule
Model & $|\mathcal{V}|$=100 & $|\mathcal{V}|$=200  \\
\midrule
GMN                       & 85.22   & 74.31    \\
GMN-HAP                   & 93.44  & 84.62    \\
HAP-MeanPool			 &57.22		&58.53	\\
HAP-MeanAttPool			&83.52		&87.84	\\
HAP-SAGPool				&59.01		&59.13	\\
HAP-DiffPool			&64.04		&59.22	\\
\textbf{HAP (ours)}                       & \textbf{98.51}  & \textbf{98.53}  \\ 
\bottomrule
\end{tabular}	
\end{table}

\section{Conclusion and Future Work}
In this paper, we introduce a novel graph pooling framework HAP for hierarchical graph-level representation learning by adaptively leveraging the graph structures. The key innovation of HAP is the graph coarsening module, assisted by novel graph pattern property extracting technique GCont and cross-level attention mechanism MOA. HAP clusters local substructures through a newly proposed cross-level attention mechanism MOA. MOA mechanism helps it to naturally focus more on close neighborhood while effectively capture higher-order dependency that may contain important information. We also propose GCont, an auto-learned global graph content that sustains the cross-attention process. HAP leverages GCont to provide global guidance in graph coarsening. It extracts graph pattern properties to make the pre- and post-coarsening graph content maintain stable without loss of significant information. The learning of GCont also facilitates generalization across graphs with the same form of features. Theoretically analysis and extensive experiments demonstrate that HAP and the key component graph coarsening module achieve state-of-the-art performance on four downstream tasks.

HAP shows its potential to improve other graph learning methods by getting a more informative graph embedding. Furthermore, there are incredible opportunities for HAP to be further extended to more complex networks such as attributed networks and heterogeneous networks which may be more common in real-world applications.

%

\bibliographystyle{IEEEtran}
\bibliography{HAP}
\vspace{-10mm}
\begin{IEEEbiography}[{\includegraphics[width=1in,height=1.25in,clip,keepaspectratio]{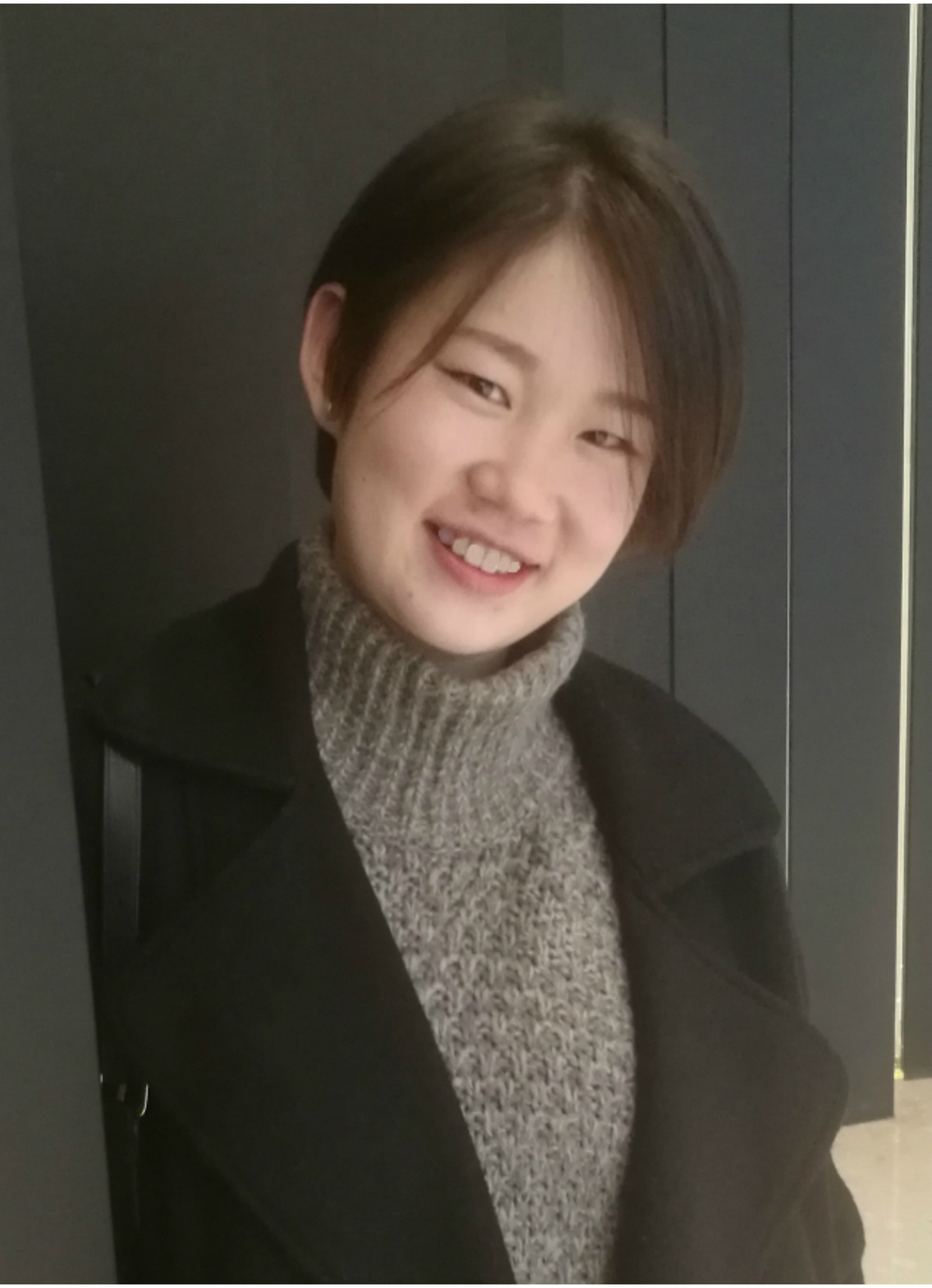}}]{Ning Liu}
is a Ph.D. student in computer science and technology from College of Computer, National University of Defense Technology, Changsha, China. Her research interests include graph representation learning, graph data analytic, and graph processing systems.
\end{IEEEbiography}
\vspace{-10mm}

\begin{IEEEbiography}[{\includegraphics[width=1in,height=1.25in,clip,keepaspectratio]{jsl.pdf}}]{Songlei Jian}
received the B.Sc. degree and Ph.D. degree in computer science from College of Computer, National University of Defense Technology, Changsha, China, in 2013 and 2019, respectively. She is currently an Assistant Research Fellow with the School of Computer, NUDT. Her research interests include representation learning, multimodal learning and anomaly detection.
\end{IEEEbiography}
\vspace{-10mm}

\begin{IEEEbiography}[{\includegraphics[width=1in,height=1.25in,clip,keepaspectratio]{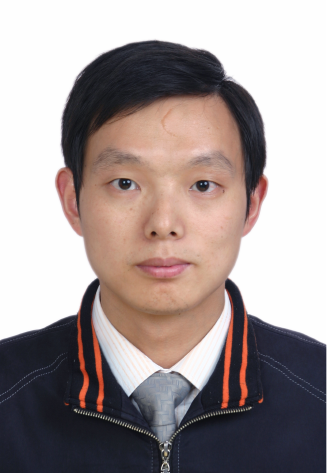}}]{Dongsheng Li}
received the B.Sc. degree (with honors) and Ph.D. degree (with honors) in computer science from College of Computer, National University of Defense Technology, Changsha, China, in 1999 and 2005, respectively. He was awarded the prize of National Excellent Doctoral Dissertation of PR China by Ministry of Education of China in 2008. He is now a full professor at National Lab for Parallel and Distributed Processing, National University of Defense Technology, China. His research interests include parallel and distributed computing, Cloud computing, and large-scale data management.
\end{IEEEbiography}
\vspace{-10mm}

\begin{IEEEbiography}[{\includegraphics[width=1in,height=1.25in,clip,keepaspectratio]{zym.pdf}}]{Yiming Zhang}
(Member, IEEE) received the B.Sc. and M.Sc. degrees in mechanics engineering and the Ph.D. degree in computer science from the National University of Defense Technology (NUDT), Changsha, Hunan, China, in 2001, 2003, and 2008, respectively. He is currently an Associate Professor with the School of Computer, NUDT. He is an associate editor of IEEE Transactions on Services Computing. His current research interests include operating systems, networking, and distributed storage. He received the China Computer Federation (CCF) Distinguished Ph.D. Dissertation Award in 2011.
\end{IEEEbiography}
\vspace{-10mm}

\begin{IEEEbiography}[{\includegraphics[width=1in,height=1.25in,clip,keepaspectratio]{lzq.pdf}}]{Zhiquan Lai}
received his Ph.D, M.S. and B.S. degrees in Computer Science from National University of Defense Technology (NUDT) in 2015, 2010 and 2008 respectively.
He is currently an assistant professor in the National Key Laboratory for Parallel and Distributed Processing of NUDT.
He worked as a research assistant at Department of Computer Science, the University of Hong Kong during Oct. 2012 to Oct. 2013. His current research interests include high-performance system software, distributed machine learning, and power-aware computing.
\end{IEEEbiography}
\vspace{-10mm}

\begin{IEEEbiography}[{\includegraphics[width=1in,height=1.25in,clip,keepaspectratio]{xhz.pdf}}]{Hongzuo Xu}
received the bachelor's and master's degree from the National University of Defense Technology, China in 2017 and 2019. He is currently pursuing his Ph.D. degree in the College of Computer, National University of Defense Technology. His research interests include anomaly detection, weakly-supervised learning and data mining. He is a student member of the IEEE.
\end{IEEEbiography}
%
%




\end{document}